\documentclass[journal]{IEEEtran}
\usepackage[pagebackref,breaklinks,colorlinks]{hyperref}
\usepackage{url}
\usepackage[utf8]{inputenc} % allow utf-8 input
\usepackage[T1]{fontenc}    % use 8-bit T1 fonts
\usepackage{hyperref}       % hyperlinks
\usepackage{url}            % simple URL typesetting
\usepackage{booktabs}       % professional-quality tables
\usepackage{amsfonts}       % blackboard math symbols
\usepackage{nicefrac}       % compact symbols for 1/2, etc.
\usepackage{microtype}      % microtypography
\usepackage{xcolor}         % colors
\usepackage{amsmath}
\usepackage{multicol}
\usepackage{multirow}
\usepackage{graphicx}
\usepackage{wrapfig}
\usepackage{dsfont}
\usepackage{textcomp}
\usepackage{xcolor}
\usepackage{bm}
\usepackage{bbm}
\usepackage[varbb]{newpxmath}
\usepackage{algorithm}
\usepackage{algorithmic} 
\usepackage{multirow}
\usepackage{caption}
\usepackage{subcaption}
\usepackage{supertabular}
\usepackage{mathrsfs}
\usepackage{amsmath}

\begin{document}

% \title{A Comparative Evaluation of Recent Universal Adversarial Perturbations\\

\title{Exploring Frequencies via Feature Mixing and Meta-Learning for Improving Adversarial Transferability}

\author{
        Juanjuan Weng,
        Zhiming Luo,
        Shaozi Li
        % Dazhen Lin,
        % Zhun Zhong
\IEEEcompsocitemizethanks{
% \IEEEcompsocthanksitem This work is supported by the National Natural Science Foundation of China (No. 62276221); the Natural Science Foundation of Fujian Province of China (No. 2022J01002); EU H2020 project AI4Media (No. 951911).
\IEEEcompsocthanksitem  J. Weng, Z. Luo (Corresponding author), and S. Li are with the Department of Artificial Intelligence, Xiamen University,
Xiamen 361005, China.
% \IEEEcompsocthanksitem Z. Zhong is with the School of Computer Science,
% University of Nottingham, Nottingham NG7 2RD, United Kingdom.
}
}

\maketitle
\begin{abstract}
Recent studies have shown that Deep Neural Networks (DNNs) are susceptible to adversarial attacks, with frequency-domain analysis underscoring the significance of high-frequency components in influencing model predictions. Conversely, targeting low-frequency components has been effective in enhancing attack transferability on black-box models. In this study, we introduce a frequency decomposition-based feature mixing method to exploit these frequency characteristics in both clean and adversarial samples. Our findings suggest that incorporating features of clean samples into adversarial features extracted from adversarial examples is more effective in attacking normally-trained models, while combining clean features with 
the adversarial features extracted from low-frequency parts decomposed from the adversarial samples yields better results in attacking defense models. However, a conflict issue arises when these two mixing approaches are employed simultaneously.
To tackle the issue, we propose a cross-frequency meta-optimization approach comprising the meta-train step, meta-test step, and final update. In the meta-train step, we leverage the low-frequency components of adversarial samples to boost the transferability of attacks against defense models. Meanwhile, in the meta-test step, we utilize adversarial samples to stabilize gradients, thereby enhancing the attack's transferability against normally trained models. For the final update, we update
the adversarial sample based on the gradients obtained from both meta-train and meta-test steps.
Our proposed method is evaluated through extensive experiments on the ImageNet-Compatible dataset, affirming its effectiveness in improving the transferability of attacks on both normally-trained CNNs and defense models. 
    The source code is available at \textcolor{blue}{\href{ https://github.com/WJJLL/MetaSSA}{Link}}. 
\end{abstract}


\section{Introduction}
%In the past decade, Deep Neural Networks (DNNs) have seen tremendous success in various tasks. However, 
Deep Neural Networks (DNNs) are susceptible to adversarial attacks~\cite{goodfellow2015explaining,wei2023towards,agarwal2022crafting}, where adding human quasi-imperceptible perturbations can cause the model to produce incorrect predictions. Building upon the foundational work of Goodfellow et al.\cite{goodfellow2015explaining}, various attack strategies have been developed, including gradient-based~\cite{kurakin2018adversarial,dong2018boosting, lin2020nesterov, dong2019evading} and input augmentation-based~\cite{xie2019improving,lin2020nesterov,byun2022improving}. 
Furthermore, adversarial examples learned from a surrogate model can be leveraged to attack black-box models, thereby heightening security concerns. As a result, crafting adversarial samples with high transferability can reveal the vulnerabilities of current DNNs and subsequently be employed to enhance their robustness.

On the other aspect, frequency domain analysis has been utilized to investigate the DNNs. Yin et al.~\cite{yin2019fourier} observed that normally trained CNNs are highly susceptible to additive perturbations in the high-frequency domain. Furthermore, Wang et al.~\cite{wang2020high} discovered that high-frequency components play a significant role in determining the final predictions of DNNs. These observations suggest that current attack methods primarily affect the high-frequency components of the input data. Thus, some defense methods~\cite{madry2017towards,tramer2017ensemble} have been proposed to mitigate the influence of high-frequency perturbations. Nonetheless, several frequency-based adversarial attacks with higher transferability have been proposed. Sharma et al.\cite{sharma2019effectiveness} and Luo et al.\cite{luo2022frequency} exclusively use low-frequency components for their attacks. Long~\cite{long2022frequency} introduced a frequency domain augmentation method to enhance transferability. In light of these discoveries, we argue that more effectively attacking the low-frequency components of inputs could further increase the success rate when targeting different defense models.

%{\color{red} Why using feature mixing}

To achieve this goal,
we have developed a feature mixing approach to better exploit the diverse frequency components present in both clean and adversarial samples. This strategy draws inspiration from the clean feature mixup (CFM)~\cite{byun2023introducing} in improving the targeted transferability. Contrasting with the CFM's reliance on original clean features for the mixup, we first employ frequency decomposition to explicitly utilize the low-frequency part and high-frequency part $(x_{l}, x_{h})$ in a clean image $x$. By doing so, we can more effectively utilize these distinct frequency parts in the feature mixing process.
Additionally, we perform two types of feature mixing operations to effectively utilize different frequency parts of the adversarial samples, namely ``Low-Frequency Adversarial Features Mixing (LF-AFM)'' and ``Adversarial Features Mixing (AFM)''. In the first type, we integrate the clean low-frequency and high-frequency features into the adversarial features extracted from low-frequency part $\tilde{x}_l$ decomposed from the adversarial example $\tilde{x}$.
The second type involves mixing these clean low-frequency and high-frequency features into the adversarial features extracted from the adversarial example $\tilde{x}$. The feature mixing is performed at a randomly selected layer, and experiments verified that both types of feature mixing can enhance the transferability of adversarial attacks. 

\begin{figure}[t]
    \centering
    \begin{subfigure}{0.49\linewidth}
         \centering
\includegraphics[width=\textwidth]{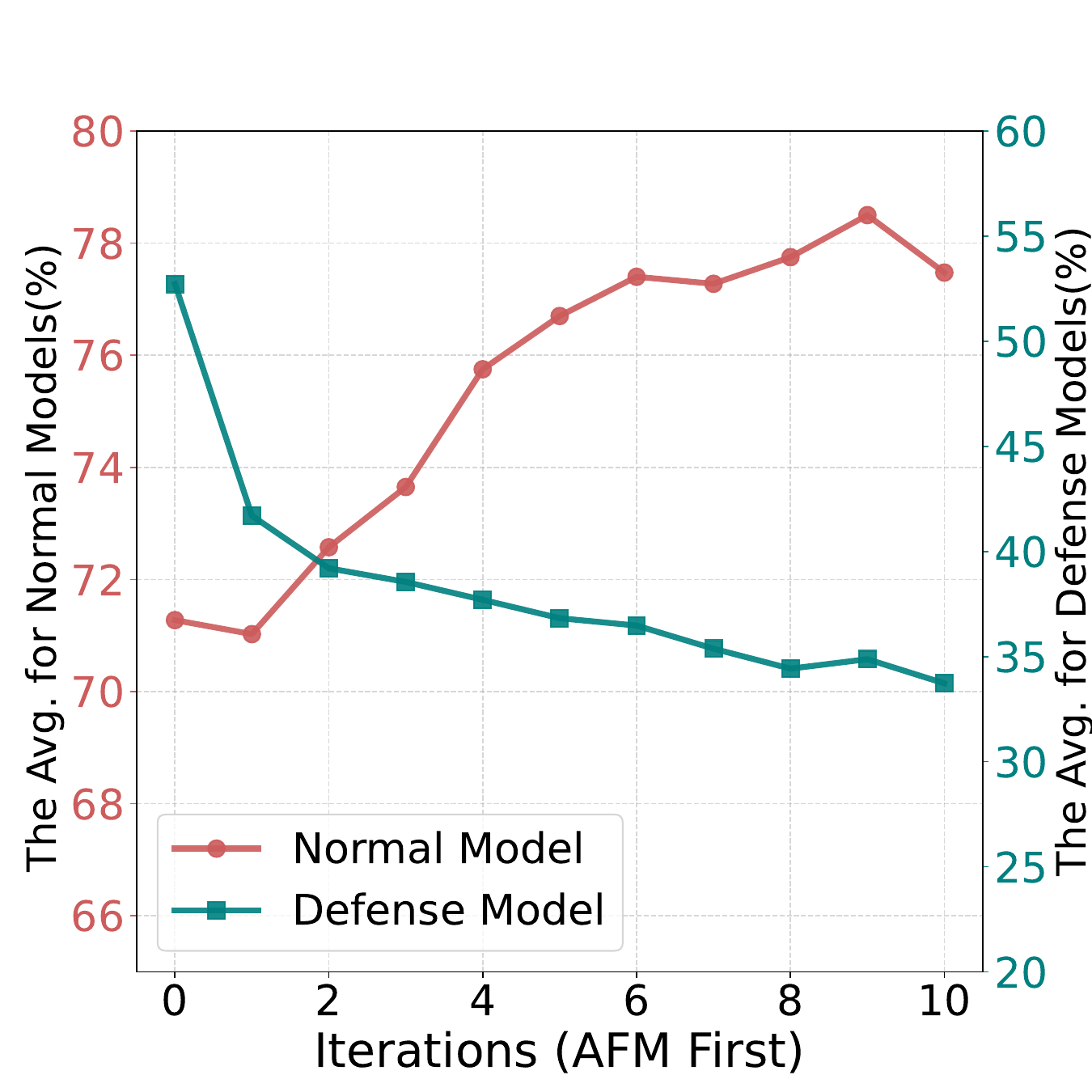}
\caption{AFM ($t$) and LF-AFM ($10-t$) on Inc-V3}  \label{fig:ince_al}
\end{subfigure}
    \begin{subfigure}{0.49\linewidth}
         \centering
\includegraphics[width=\textwidth]{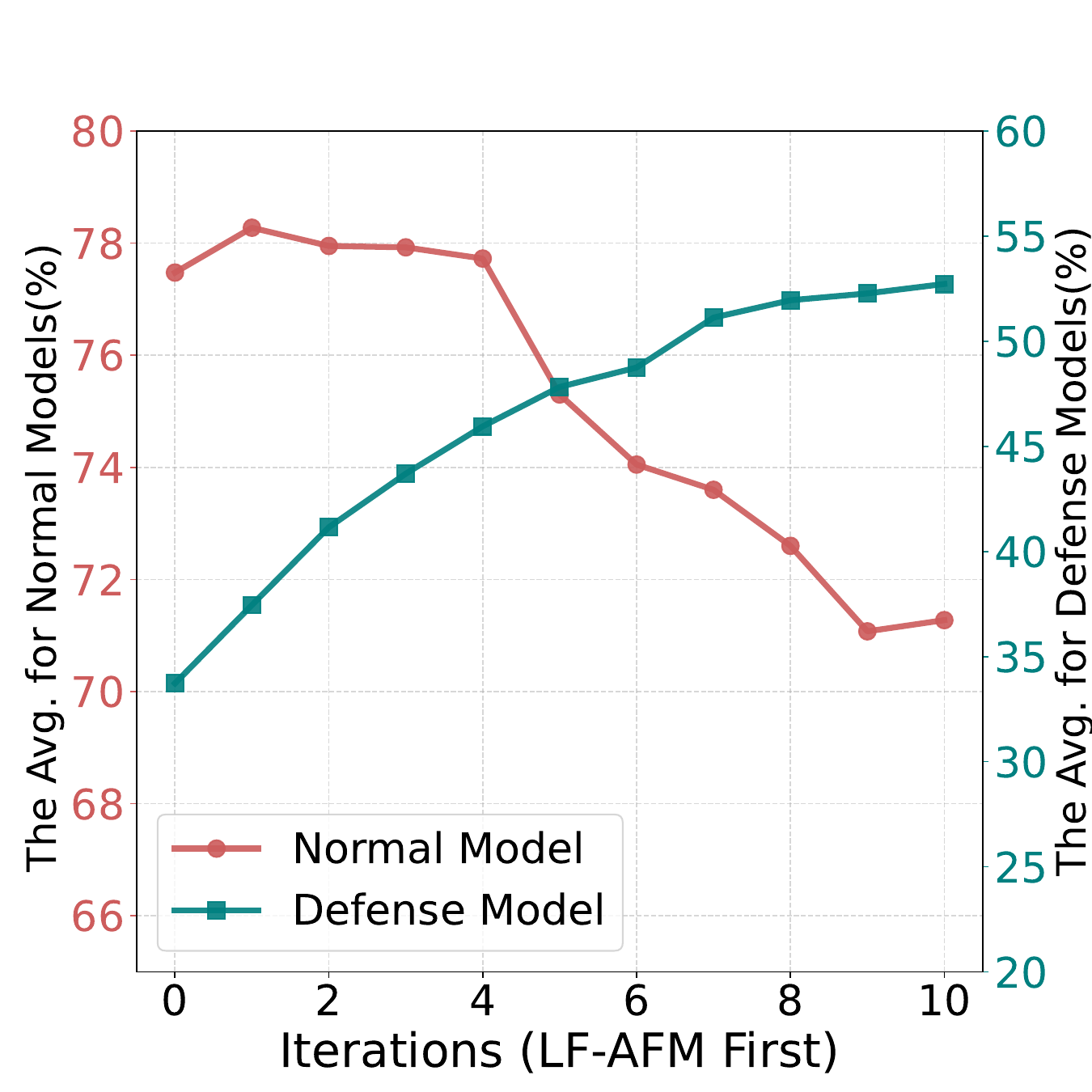}
\caption{{LF-AFM ($t$) and AFM ($10-t$) on Inc-v3}}  \label{fig:ince_la}
\end{subfigure}
    \begin{subfigure}{0.49\linewidth}
         \centering
\includegraphics[width=\textwidth]{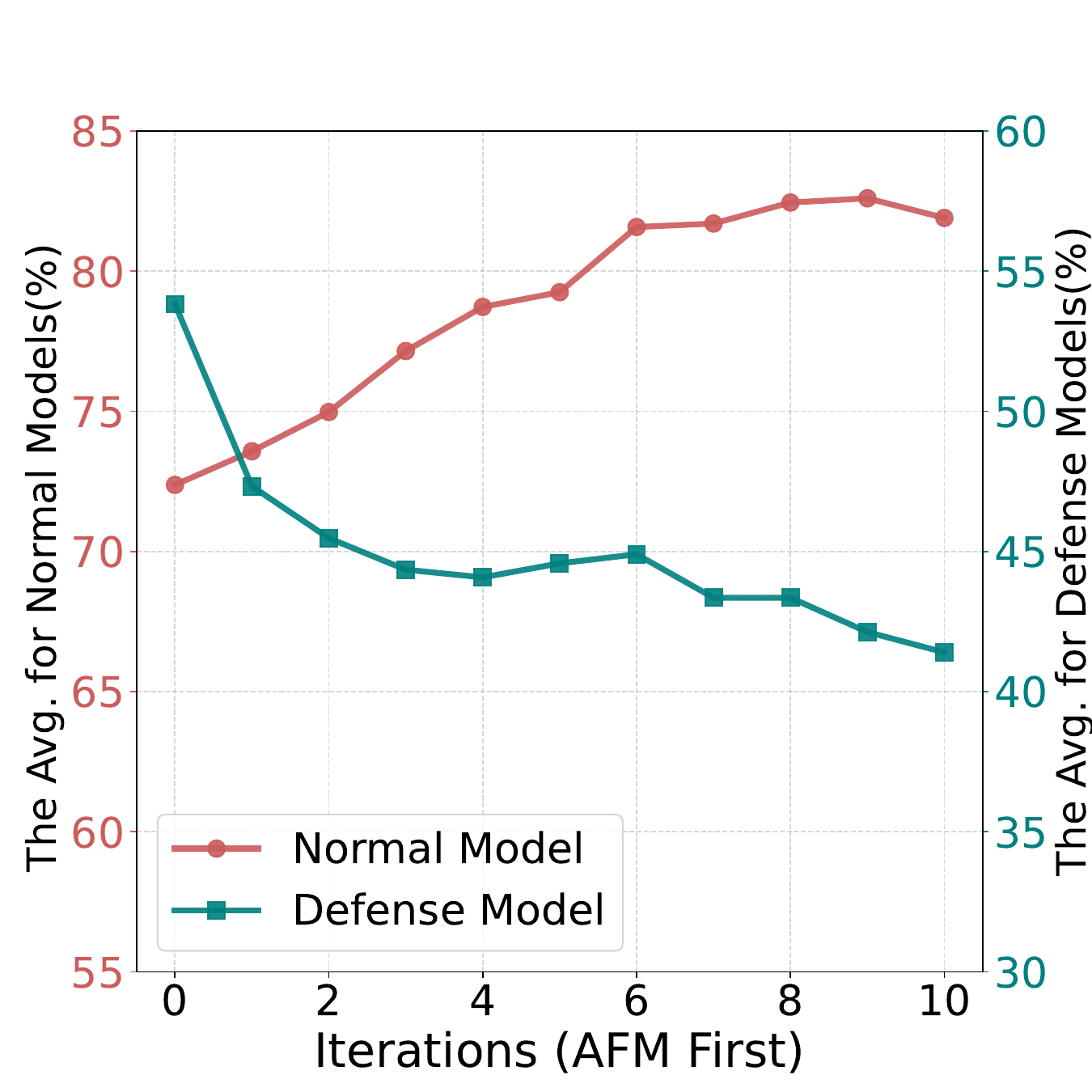}
\caption{AFM ($t$) and LF-AFM ($10-t$) on Res-101}  \label{fig:res_al}
\end{subfigure}
    \begin{subfigure}{0.49\linewidth}
         \centering
\includegraphics[width=\textwidth]{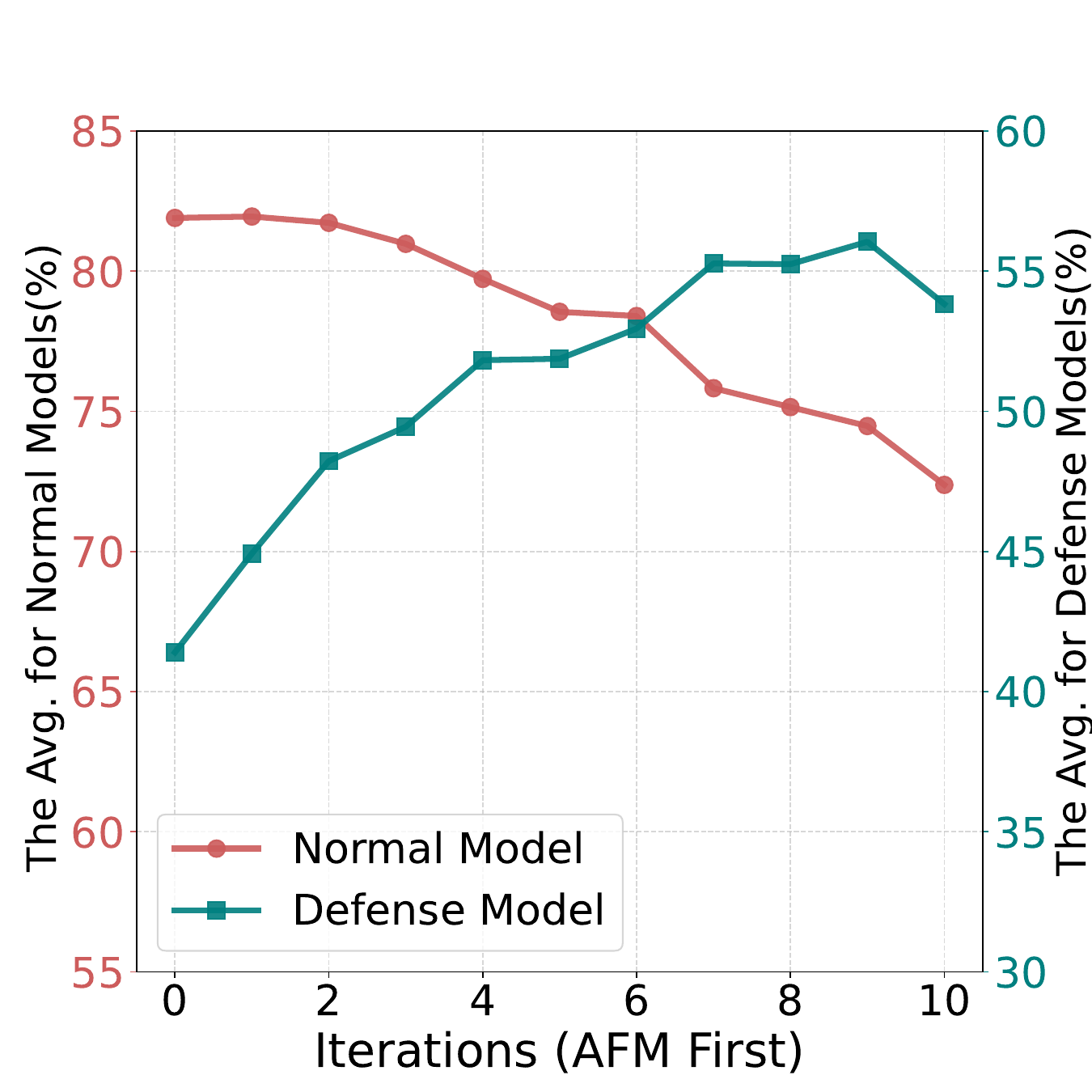}
\caption{LF-AFM ($t$) and AFM ($10-t$) on Res-101
}  
\label{fig:res_la}
\end{subfigure}
    \caption{The average success rates (\%) of adversarial attacks are evaluated on four normally trained models and four defense models. Adversarial examples are crafted by employing both AFM and LF-AFM simultaneously during attack iterations using the MI-FGSM.  ``AFM ($t$) and LF-AFM ($10-t$) on Inc-v3'' indicates the use of AFM for the initial $t$ iterations, followed by LF-AFM for the remaining $10-t$ iterations with Inc-v3 regarded as the source model. }
\label{fig:conflict}
\end{figure}

However, our further observations reveal that LF-AFM proves more effective in attacking defense models, while AFM is preferable for attacking normally-trained models. Additionally, as detailed in Figure~\ref{fig:conflict} (refers to Section~\ref{sec:conflict}), \textbf{the simultaneous use of both AFM and LF-AFM during attack iterations will lead to a conflict}: employing AFM with more iterations results in a steady increase in average attack success rates against normally-trained models but a decrease in attack against adversarially trained models. Conversely, LF-AFM yields opposite effects.

%Building upon the previous findings, we investigate the impact of using original samples and their low-frequency counterparts to craft adversarial examples. Initially updating perturbations based on the original samples for several iterations and subsequently switching to low-frequency parts for optimization results in a significant improvement in attacking defense models. However, this approach leads to a decline in the success rate of attacking regular CNNs. In a reversed optimization order, attacking regular CNNs will show an increase in success rate while defense models will exhibit a decrease. Thus, it will lead to a conflict issue.

Meta-learning, a concept focused on learning how to learn, has been employed in adversarial attacks~\cite{yuan2021meta} to aggregate gradients from various source models. %Inspired by this, our motivation is to enhance gradient stabilization between two tasks. 
To address the aforementioned conflicts, we further propose a cross-frequency meta-optimization to enhance the transferability for attacking both normally trained CNNs and defense models. 
% decompose the original input image $x$ and its corresponding adversarial sample $\tilde{x}$ into low-frequency part and high-frequency part based on the Discrete Wavelet Transform (DWT), denoted as $(x_{l}, x_{h})$ and $(\tilde{x}_{l}, \tilde{x}_{h})$. Subsequently, we perform two types of feature mixing operations at a randomly selected layer of the surrogate CNN. The first one is mixing features of $(x_{low}, x_{high}, x'_{low})$. The other one is mixing between $(x_{low}, x_{high}, x')$. 
%Finally, we introduce a cross-frequency meta-learning framework to optimize the adversarial examples. 
In the meta-train step, we iteratively update the adversarial example based on LF-AFM, which can significantly increase the transferability of attacking defense models. In the meta-test step, we use AFM to calculate the average gradient across all intermediate adversarial samples generated in the meta-train phase. This step aims to stabilize the gradient, thus enhancing the attack's transferability against normally trained models. Lastly, we update the adversarial examples based on the gradients obtained from both meta-train and meta-test steps.

The main contributions are summarized as follows:
\begin{itemize}
    \item 
    % We introduce two novel feature mixing strategies that utilize frequency decomposition to effectively exploit both low-frequency and high-frequency components in clean and adversarial samples, which include LF-AFM and AFM for improving attacking normally-trained and defense models.
   We introduce feature mixing based on frequency decomposition and cross-frequency meta-optimization for learning adversarial examples with high transferability. 
    
    \item Our proposed cross-frequency meta-optimization framework can effectively resolve the conflict issue by utilizing adversarial samples and their low-frequency counterparts in the optimization process for learning the final adversarial examples.
    
    \item Extensive experimental results on the ImageNet-Compatible dataset demonstrate that the adversarial examples generated by our method exhibit higher transferability when attacking both normally trained and defense models. Especially, we can observe a significant boost in attacking various defense models.
\end{itemize}

\section{Related Works}
\subsection{Adversarial Attacks}
%Since the initial finding of adversarial samples by Szegedy et al.~\cite{szegedy2014intriguing}, various attack methods have been developed to create more disruptive adversarial perturbations and enhance the transferability of attacking black-box models, which can be primarily grouped into the following two categories: \textbf{Gradient-based} and \textbf{Input augmentation-based}.
Recently proposed representative adversarial attack methods for enhancing transferability can be grouped into the following two categories: \textbf{Gradient-based} and \textbf{Input augmentation-based}.

\textbf{Gradient-based methods} increase the transferability of adversarial samples based on more advanced optimization techniques. %The pioneering FGSM~\cite{goodfellow2015explaining} directly generates the adversarial perturbation by maximizing the cross-entropy loss function. 
Kurakin et al.~\cite{kurakin2018adversarial} proposed the I-FGSM, which iteratively updates the perturbation with a small step based on the sign of the gradient. Following I-FGSM, several improvements have been made. For example, MI-FGSM~\cite{dong2018boosting} uses a momentum term to stabilize the gradient based on the previous iteration. Similarly, NI-FGSM uses the Nesterov accelerated gradient to accumulate the gradients over training iterations. Translation-Invarial (TI)~\cite{dong2019evading} convolves the gradient with a pre-defined kernel for mincing an ensemble of shifted inputs. Variance Tuning (VT)~\cite{wang2021enhancing} introduces
neighborhood information of the input at the last iteration to
correct the current update direction. GRA~\cite{zhu2023boosting} introduces a gradient relevance framework, which establishes the gradient relevance between
the input and its neighborhood at each iteration.

\textbf{Input augmentation-based methods} conduct input-level data augmentation for learning the adversarial perturbations. For example, the Diverse Input (DI)~\cite{xie2019improving} randomly resizes the input image and adds padding during each iteration. 
The Scale-Invariant (SI)~\cite{lin2020nesterov} augments the input image with multiple scale copies $\left(S_i(x)=x/2^i\right)$ for learning the perturbations. Admix~\cite{wang2021admix} mixes the input image with other images randomly selected in the same
batch to augment the input and then update with gradients
calculated on the mixed image. 
The gradient-based methods can be used jointly with these input augmentation-based methods to further boost the adversarial transferability.

\subsection{Frequency-based Attacks}
%Frequency domain analysis has been used to examine the robustness of deep neural networks (DNNs) in handling different frequency characteristics during the network inference process. Wang et al.~\cite{wang2020high} discovered that CNN predictions heavily rely on imperceptible high-frequency components of an image, even the low-frequency components appear identical to the original image.
%Yin et al.~\cite{yin2019fourier} find that normally trained CNNs exhibit a high susceptibility to additive perturbations in high frequencies. Meanwhile, they found that incorporating Gaussian data augmentation into the adversarial training can greatly enhance the model's ability to withstand high-frequency noises.

Several adversarial attacks~\cite{sharma2019effectiveness,duan2021advdrop,long2022frequency} have been proposed based on the frequency domain. Sharma et al.~\cite{sharma2019effectiveness} utilized the discrete cosine transform (DCT) to confine perturbations, ensuring that only the low-frequency components of the input are altered.
Duan et al.~\cite{duan2021advdrop} proposed AdvDrop, which generates adversarial examples by removing existing details from clean images in the frequency domain. Long et al.~\cite{long2022frequency} proposed a frequency domain data augmentation utilizing a random spectrum transformation, which involves two components: adding Gaussian noise and applying a Hadamard product with a Gaussian-distributed mask after the DCT. Wang et al.~\cite{wang2023lfaa} introduced the Low-Frequency Adversarial Attack (LFAA) to enhance the transferability of targeted attacks. This method trains a conditional generator to craft adversarial frequency domain information aligned with the target label. Subsequently, this information is incorporated into the low-frequency component of the input image to generate adversarial examples.
In contrast to the aforementioned methods, we observed that low-frequency adversarial features are more effective in attacking defense models, while adversarial features are preferable for attacking normally-trained models. To effectively address the conflict issue and leverage the benefits of both types of features, we
employ a frequency decomposition-based feature mixing technique and a meta-learning framework.

\subsection{Adversarial Defenses}
To mitigate the impact of adversarial samples, various defense methods have been proposed, which can be primarily categorized into two groups: \textbf{adversarial training-based methods} and \textbf{input pre-processing-based methods}. The adversarial training-based methods~\cite{madry2017towards} enhance the robustness of CNNs by incorporating both clean samples and their adversarial counterparts during training. To further increase the robustness against unknown black-box attacks, the Ensemble Adversarial Training~\cite{tramer2017ensemble} further leverages the adversarial examples generated on other static pre-trained models during training. On the other hand, the input pre-processing-based methods primarily concentrate on reducing the impact of adversarial perturbations prior to inputting images into CNNs. The representative methods are JPEG compression~\cite{guo2017countering}, random resizing and padding (R\&P)~\cite{xie2017mitigating}, and high-level representation guided denoiser (HGD)~\cite{liao2018defense}. Furthermore, there have been attempts to combine the advantages of these two types of methods to further improve defense performance, such as NeurIPS-r3 solution~\cite{thomas2017defense} and Neural Representation Purifier (NRP)~\cite{naseer2020self}. In this study, we leverage these defense mechanisms primarily to assess the disruptive nature of our proposed frequency-based adversarial attack, which can significantly increase the attack success rate against these defense methods.

\begin{figure*}[t]
    \centering    \includegraphics[width=0.95\linewidth]{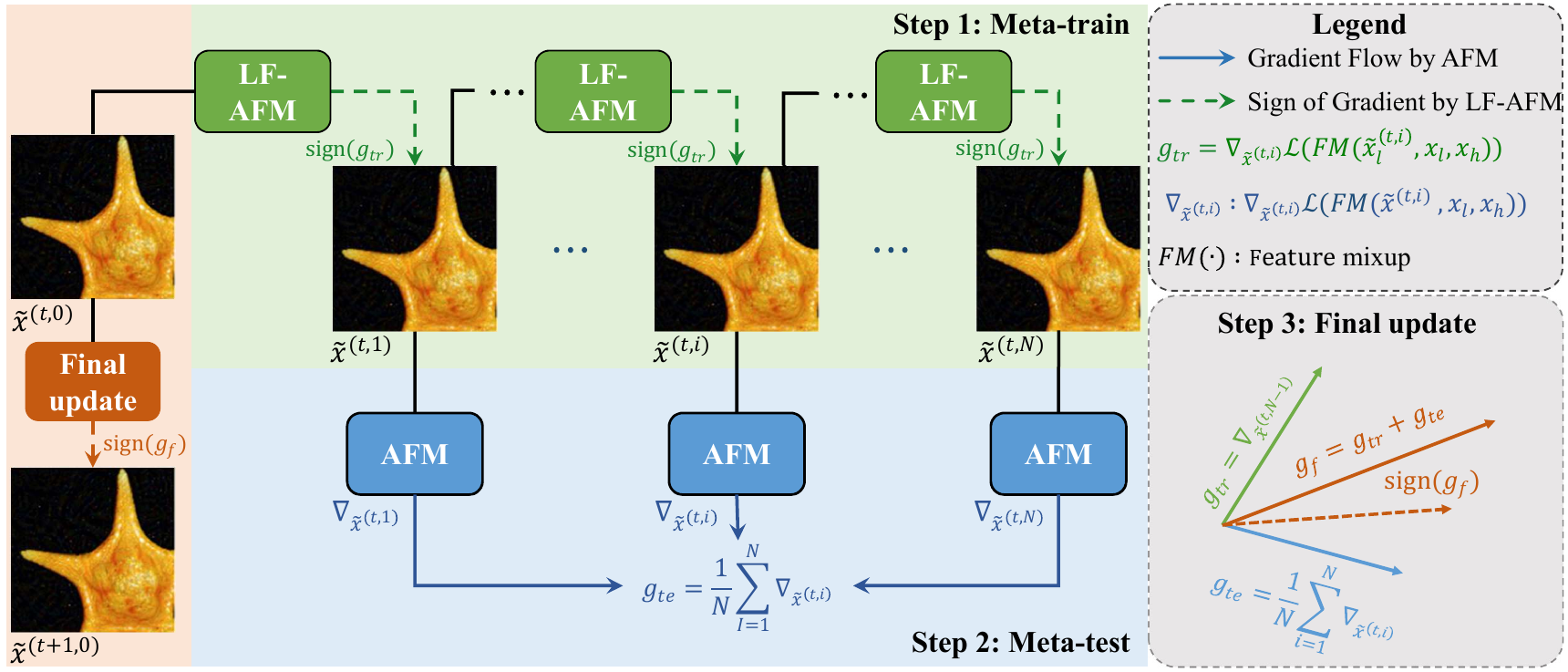}
    \caption{
The cross-frequency meta-optimization process at iteration $t$ comprises three steps: (1) low-frequency guided meta-train, (2) adversarial sample guided meta-test, and (3) final update. In the meta-train step, we sequentially use the Low-Frequency Adversarial Features Mixing (LF-AFM) to compute the current meta-train gradient $g_{tr}$ for crafting temporary adversarial examples $\tilde{x}^{(t, i)}$ at each inner iteration $i$. In the meta-test step, we average the gradients $\nabla_{\tilde{x}^{t,i}}$ using the Adversarial Features Mixing (AFM) across all intermediate $\tilde{x}^{(t, i)}$ generated in the meta-train phase. Finally, for the final update, we craft the adversarial examples $\tilde{x}^{(t+1,0)}$ based on the gradients obtained from both meta-train and meta-test steps.
}
    \label{fig:framework}
\end{figure*}

\section{Method}

\subsection{Frequency-based attack}
\subsubsection{Frequency decomposition}

From a frequency domain perspective, the high-frequency components representing noise and textures are more imperceptible than the low-frequency components containing basic object structure. Following previous study~\cite{luo2022frequency}, we also use the Discrete Wavelet Transform (DWT) to decompose the input image $x$ into four components, including one low-frequency and three high-frequency. The computation of DWT can be denoted as:
\begin{align}
    &x_{ll}=\bm{L}x\bm{L}^T,~x_{lh}=\bm{H}x\bm{L}^T, \\
    &x_{hl}=\bm{L}x\bm{H}^T,~x_{hh}=\bm{H}x\bm{H}^T,
\end{align}
where $\bm{L}$ and $\bm{H}$ are the low-pass and high-pass filters of orthogonal wavelet, respectively.

Inverse discrete wavelet transform (IDWT) typically utilizes all four components to reconstruct the image. To obtain the low-frequency part of an image, we discard the high-frequency components and reconstruct it solely using the low-frequency component as in ~\cite{luo2022frequency}, denoted as:
\begin{equation}
    x_{l} = \bm{L}^Tx_{ll}\bm{L}.
\label{eq:low}
\end{equation}
The corresponding high-frequency part can be computed by subtracting the $x_{l}$ from the original $x$: 
\begin{equation}
    x_{h} = x - x_{l}.
\label{eq:high}
\end{equation}

\subsubsection{Feature Mixing}
To improve the transferability of adversarial examples across different types of models,
we execute two types of feature mixing operations: ``Low-Frequency Adversarial Features Mixing (LF-AFM)'' and ``Adversarial Features Mixing (AFM)''. 
To accelerate feature mixing in model inferences, we first store the clean features in memory. We decompose $x$ into low-frequency and high-frequency parts, denoted as $(x_{l}, x_{h})$. Subsequently, we input $x_l$ and $x_h$ into the surrogate model $f$, storing corresponding features for each layer $\left\{ f^{1}_{x_l}, f^{2}_{x_l}, \ldots, f^{K}_{x_l}\right\}$ and $\left\{ f^{1}_{x_h}, f^{2}_{x_h}, \ldots, f^{K}_{x_h}\right\}$, where $K$ represents the number of layers. The notation $f^K_*$ denotes the feature at  $K_{th}$ layer of input $*$.

For LF-AFM, we decompose adversarial example $\tilde{x}$ to obtain the low-frequency input $\tilde{x}_l$. We then integrate the stored clean low-frequency feature $f_{x_l}^k$ and high-frequency feature $f_{x_h}^k$ into the
adversarial features extracted from low-frequency part $\tilde{x}_l$ at corresponding $k_{th}$ layer of the model $f$. The $k$ is randomly selected. Specifically, the mixed feature is computed as:
\begin{equation}
    \hat{f}^k_{
    \tilde{x}_{l}} = FM(\tilde{x}_l,x_l,x_h) = \alpha f^k_{x_{l}} + \beta f^k_{x_{h}} + \gamma f^k_{\tilde{x}_{l}},
\label{eq:mix_low}
\end{equation}
where $\alpha$, $\beta$, and $\gamma$ are three random weights, subject to the constraint $\alpha + \beta + \gamma =1$, $FM(\cdot)$ represents the mixup function at a randomly selected layer $k$.

For AFM, we mix these
clean low-frequency and high-frequency features into the
adversarial features extracted from $\tilde{x}$ at randomly selected
layers of the model $f$ in the same manner,
\begin{equation}
    \hat{f}^k_{\tilde{x}} = FM(\tilde{x},x_l,x_h) = \alpha f^k_{x_{l}} + \beta f^k_{x_{h}} + \gamma f^k_{\tilde{x}}.
\label{eq:mix_adv}
\end{equation}

Finally, the mixed features $\hat{f}^k_{\tilde{x}_{l}}$ and $\hat{f}^k_{\tilde{x}}$ will be forwarded to the subsequent layers of model $f$ to obtain the final output, respectively.

% \begin{table*}[]
% \centering
% \caption{The attack success rates (\%) on seven models by a single attack. The adversarial examples are generated on Inc-v3. }
% \label{tab:conflict}
% \resizebox{0.9\linewidth}{!}{
% \begin{tabular}{lllllllll}
% \bottomrule[1.2pt]
% \multicolumn{2}{l}{\textbf{Source: Inc-v3}} & \multicolumn{7}{c}{Target Model} \\
% \cline{2-9}
% Strategy  & Inc-v3 & Inc-v4 & IncRes-v2 & Res-101 & Inc-v3$_{adv}$ & Inc-v3$_{ens3}$ & Inc-v3$_{ens4}$ & IncRes-v2$_{ens}$ \\
% \hline
% AFM & 99.5 & 76.1 & 73.3 & 65.9 & 40.2 & 38.2 & 37.2 & 19.2 \\
% LFM & 98.8 & 66.9 & 65.5 & 54.3 & 60.8 & 52.7 & 57.3 & 40.1 \\
% AFM, LF-AFM & 99.0 & 72.0 & 69.9 & 60.1 & 56.4 & 50.0 & 54.5 & 34.9 \\
% LF-AFM, AFM & 98.5 & 72.8 & 70.4 & 63.8 & 41.7 & 38.1 & 39.5 & 22.3 \\
% iterative order swapping & 97.8 & 74.6 & 70.3 & 64.5 & 44.7 & 43.0 & 45.0 & 26.0 \\
% \bottomrule[1.2pt]
% \multicolumn{2}{l}{\textbf{Source: Res-101}} & \multicolumn{7}{c}{Target Model} \\
% \cline{2-9}
% Strategy  & Inc-v3 & Inc-v4 & IncRes-v2 & Res-101 & Inc-v3$_{adv}$ & Inc-v3$_{ens3}$ & Inc-v3$_{ens4}$ & IncRes-v2$_{ens}$ \\ \hline
% AFM & 79.6 & 75.5 & 74.1 & 99.5 & 46.4 & 46.3 & 43.8 & 29.5 \\
% LFM & 68.2 & 61.8 & 62.3 & 99.4 & 60.7 & 56.3 & 55.9 & 45.2 \\
% AFM, LF-AFM & 76.9 & 73.3 & 70.6 & 99.2 & 48.8 & 47.8 & 46.5 & 31.7 \\
% LF-AFM, AFM & 75.2 & 71.0 & 70.2 & 99.8 & 58.9 & 54.9 & 56.2 & 42.2 \\
% iterative order swapping & 77.6 & 73.2 & 72.8 & 99.6 & 50.3 & 50.8 & 50.8 & 35.1\\
% \bottomrule[1.2pt]
% \end{tabular}}
% \end{table*}

\subsubsection{Conflict issue}
\label{sec:conflict}

We simultaneously employ both AFM and LF-AFM during attack iterations based on the MI-FGSM to craft adversarial examples. Subsequently, we evaluate the average attack success rates on four normally-trained models (Inc-v3, Inc-v4, IncRes-v2, and Res-101) and four defense models (Inc-v3$_{adv}$, Inc-v3$_{ens3}$, Inc-v3$_{ens4}$, and IncRes-v2$_{ens}$). Specifically, as shown in Figure~\ref{fig:ince_al}, AFM is used for the initial $t$ iterations, succeeded by LF-AFM for the remaining $10-t$ iterations with Inc-v3 regarded as the source model. In contrast, Figure~\ref{fig:ince_la} shows the results of LF-AFM for the initial $t$ iterations followed by AFM for the remaining $10-t$ iterations. Additionally, Figure~\ref{fig:res_al} and Figure~\ref{fig:res_la} illustrate the average attack success rates of Res-101 as the source model.

From Figure~\ref{fig:ince_al} (AFM for the initial $t$ iterations), we can have the following observations: \textbf{1)} The average attack success rates for attacking normally-trained models consistently increase along with the increase in number of iterations using AFM. However, the attack results against adversarially trained models decrease. \textbf{2)} Only applying LF-AFM ($t=0$) across all iterations shows preferable results for attacking adversarially trained models, while with limited effectiveness of attacking normally-trained models.  \textbf{3)} Only employing AFM ($t=10$) for all iterations demonstrates high effectiveness when attacking normally trained models but limited impact on adversarially trained models.
\textbf{These phenomena highlight a conflict that arises from the joint usage of both AFM and LF-AFM during attack iterations.}

Additionally, we assess the performance of using LF-AFM in the initial $t$ iterations. As shown in in Figure~\ref{fig:ince_la}, as the utilization of LF-AFM rise, the average attack success rate on normally-trained models notably decreases when $t \ge 5$. Meanwhile, the performance against defense models shows a consistent improvement. %\textbf{This finding indicates that employing LF-AFM initially, followed by AFM for fine-tuning, is more effective than the reverse order of AFM followed by LF-AFM.}
This finding further underscores the conflicting issue when jointly use AFM and LF-AFM in an reverse order compared to Figure~\ref{fig:ince_al}. 

Finally, when using ResNet-101 as the surrogate model, as depicted in Figure~\ref{fig:res_al} and Figure~\ref{fig:res_la}, we can have similar findings as using Inc-v3 as the surrogate model (Figure~\ref{fig:ince_al} and Figure~\ref{fig:ince_la}).

% \textbf{1)} The first strategy (1st row) involves AFM for all iterations, demonstrating high effectiveness against normally trained models but limited impact on adversarially trained models. 
% \textbf{2)} The second strategy (2nd row) applies LF-AFM throughout all iterations, showing improved results against adversarially trained models, yet limited effectiveness on normally-trained models. \textbf{3)} 
% In the 3rd row, this strategy involves initially using  AFM for the first half of the iterations and then switching to LF-AFM. Compared to the 1st row, it boosts performance against defense models but reduces it against normally trained models. 
% \textbf{4)} In the 4th row, this strategy reverses this optimization order, starting with LF-AFM and then switching to AFM. Compared to the 2nd row, it increases success against normally trained models but decreases it against defense models. \textbf{5)} The fifth strategy (5th row) iteratively alternates between AFM and LF-AFM, leading to similar results with the third strategy. \textcolor{red}{These results highlight a conflict when combining both methods}.

\subsection{Cross-Frequency Meta Optimization}
% 缺少meta 的概念
%{\color{red} Meta-learning, centered around the concept of learning how to learn, has been utilized in adversarial attacks~\cite{yuan2021meta} to ensemble gradients from various source models. Inspired by this for learning ensemble attacks, our motivation is to address the conflict issue by achieving enhanced gradient stabilization between two tasks—utilizing low-frequency components to guide adversarial attacks and original adversarial samples to refine the attacks.}

To address the conflict arising from the simultaneous use of AFM and LF-AFM during attack iterations, we introduced a cross-frequency meta-optimization technique. This method aims to improve the transferability of attacks across both normally-trained CNNs and defense models.
%In this step, we leverage meta-learning to address the previously discussed conflict issue when using AFM and LF-AFM jointly and propose a cross-frequency meta-optimization procedure. 
As depicted in Figure~\ref{fig:framework}, the overall optimization procedure mainly consists of the following three steps. Given the adversarial sample $\tilde{x}^t$ at the iteration $t$, we first use its corresponding low-frequency component to perform Low-Frequency Adversarial Features Mixing (LF-AFM) for the meta-train step. Then, we use the intermediate adversarial samples obtained from the meta-train step to conduct Adversarial Features Mixing (AFM) for the meta-test step, stabilizing the gradient to handle the conflict issue as described in Section~\ref{sec:conflict}. 
Finally, we update the adversarial sample for this iteration using gradients from both the meta-train and meta-test steps.
% For better readability, we will omit the subscript $t$ in $\tilde{x}_t$ in the following sections.

\subsubsection{Low-frequency guided meta-train}

The meta-train step conducts a sequential optimization process based on the low-frequency part $\tilde{x}^{t}_l$ of the adversarial sample $\tilde{x}^{t}$. In each inner iteration, we use the $\tilde{x}^{t}$ to initialize $\tilde{x}^{(t,0)}$. Besides, as shown in Figure~\ref{fig:ince_la} and Figure~\ref{fig:res_la}, there is sligt decrease in the average success rates of attacks on normal models while experiencing a significant improvement for attacks on defense models when $t \leq 4$. Therefore for each inner iteration $i$, we first employ the LF-AFM to obtain the final output. Then, we compute the meta-train gradient based on the low-frequency branch, as follows:
\begin{equation}
    g_{tr} = \nabla_{\tilde{x}^{(t,i)}} \mathcal{L}(FM(\tilde{x}^{(t,i)}_l, x_l,x_h)\big),
\label{eq:meta-train}
\end{equation}
where $\mathcal{L}$ is loss which is usually the cross-entropy loss. 
Finally, we update the adversarial examples $\tilde{x}^{(t,i)}$ using the MI-FGSM method to obtain temporary adversarial example $\tilde{x}^{(t,i+1)}$, represented by: 
\begin{equation}
 \tilde{x}^{(t,i+1)} \leftarrow \Pi \{ \tilde{x}^{(t,i)} + \alpha \cdot \text{sign}( g_{tr})\},
\label{eq:meta-adv}
\end{equation}
where $\alpha$ denotes the step size, $\Pi \{ \cdot \}$  denotes the projection function ensured by $l_p$ constrain.
After optimizing for $N$ consecutive inner steps, we obtain the intermediate adversarial examples $\tilde{\mathcal{X}} =\{ \tilde{x}^{(t,1)},\tilde{x}^{(t,2)}\cdots,\tilde{x}^{(t,N)}  \}$, which will be used for computing the stabilization gradient in the following meta-test step. 

% Besides, we set the gradient of this meta-train step, $g_{tr}$, equal to the gradient at the final iteration:
% \begin{equation}
%     g_{tr} = \nabla_{\tilde{x}^{(t,N-1)}} \mathcal{L}(FM(\tilde{x}^{(t,N-1)}_l)\big).
% \label{eq:meta-train}
% \end{equation}

%Then N adversarial perturbations, can be crafted iteratively by consecutive steps updating with Eq.
\begin{algorithm}[tb]
    \caption{The proposed attacking method}
    \label{alg:CFMO}
    \textbf{Input}: A clean input image $x$; a source model $f$; iterations $T$; sample quantity $N$.\\
    \textbf{Output}: The adversarial sample $\tilde{x}$ 
    
    \begin{algorithmic}[1] %[1] enables line numbers
    \STATE Obtain the low-frequency and high-frequency images ($x_l$, $x_h$) from the clean input $x$ using Eq.\ref{eq:low} and Eq.\ref{eq:high}.
    \STATE Storing clean features of $x_l$ and $x_h$ for each layer by $f$
        \STATE $\alpha=\epsilon / T$, $\tilde{x}^{0} = x$
        \FOR{$t=1 \rightarrow T$}
        \STATE $\tilde{x}^{(t,0)} = \tilde{x}^{t}$,  $\tilde{\mathcal{X}} = \emptyset $
            \FOR{$i=1 \rightarrow N$}
                \STATE Perform the feature mixing for low-frequency input $\tilde{x}^{(t,i)}_l$ by Eq.~\ref{eq:mix_low} then obtain the final output 
                \STATE Compute meta-train gradient 
                $g_{tr}$ by Eq.~\ref{eq:meta-train}
                \STATE
            Obtain temp $\tilde{x}^{(t,i+1)}_l$ by Eq.\ref{eq:meta-adv} and append it to $\tilde{\mathcal{X}}$
            \ENDFOR
            \STATE Perform feature mixing for each $\tilde{x}^{(t,i)}$ in $\tilde{\mathcal{X}}$ by Eq.~\ref{eq:mix_adv} then obtain the final output
            \STATE Compute the meta-test gradient based on $\tilde{\mathcal{X}}$ by Eq.~\ref{eq:meta-test}
            \STATE Update the adversarial 
sample $\tilde{x}^{t}$ to obtain $\tilde{x}^{t+1}$  by Eq.~\ref{eq:adv}
        \ENDFOR
        \STATE \textbf{return} $\tilde{x}^{(T)}$
    \end{algorithmic}
\end{algorithm}

\subsubsection{Adversarial sample guided meta-test}

In the previous meta-train step, when only updating the adversarial examples based on low-frequency adversarial features mixing, it will introduce a bias issue toward attacking adversarially-trained models. We further conduct a meta-test step to stabilize the gradients to increase the transferability of attacking regular models. In this meta-test step, we use the intermediate adversarial samples to obtain the gradient instead of using the low-frequency parts. 

Initially, we compute the gradients concerning each intermediate adversarial sample $\tilde{x}^{(t, i)}$ while simultaneously conducting adversarial feature mixing to enhance the transferability across normally trained models. Subsequently, we calculate the average gradient from all intermediate adversarial samples to obtain the gradient for the current meta-test step. This procedure can be summarized as follows:
\begin{equation}
   g_{te} = \frac{1}{N} \sum_{i=1}^{N} \nabla_{\tilde{x}^{(t,i)}} \mathcal{L}(FM(\tilde{x}^{(t,i)},x_l,x_h)\big).
\label{eq:meta-test}
\end{equation}

\subsubsection{Final update}

After obtaining the gradients $g_{tr}$ and $g_{te}$ from the previous meta-train and meta-test steps, we update the adversarial sample by combining these two gradients. The computation is denoted as:
\begin{equation}
    \tilde{x}^{t+1} \leftarrow \Pi \{\tilde{x}^t + \alpha \cdot \text{sign}(g_{tr} + g_{te})\}.
\label{eq:adv}
\end{equation}
where $\alpha=\epsilon/T$ ensures that $\tilde{x}$ remains within the $l_p$-norm when optimized for a maximum of $T$ iterations. 

To summarize, the whole training procedure of our method is described in Alg.~\ref{alg:CFMO}.

\section{Experiments}

\begin{table*}[]
\centering
\caption{The attack success rates (\%) on seven models by a single attack. The adversarial examples are generated on Inc-v3, Inc-v4,
IncRes-v2, and Res-101 separately. $*$ denotes the success rate of the white-box attack and the result in bold is the best.}
\label{tab:single}
% \resizebox{0.95\linewidth}{!}{
\begin{tabular}{lcccccccc}
\bottomrule[1.2pt]
\textbf{Source: Inc-v3} & \multicolumn{7}{c}{Target model} & \multicolumn{1}{l}{} \\
 \cline{2-8}
Attack & Inc-v3 & Inc-v4 & IncRes-v2 & Res-101 & Inc-v3$_{ens3}$ & Inc-v3$_{ens4}$ & IncRes-v2$_{ens}$ & Avg. \\ \hline

SINI & \textbf{100.0*} & 74.0 & 73.1 & 64.2 & 35.1 & 33.9 & 18.0 & 56.9 \\
VTMI & 99.9* & 77.2 & 72.2 & 64.8 & 41.1 & 40.8 & 24.3 & 60.0 \\
Admix & \textbf{100.0*} & 82.7 & 78.8 & 75.7 & 44.3 & 42.8 & 24.5 & 64.1 \\
GRA & 99.9* & 86.5 & 84.0 & 79.2 & 59.9 & 59.6 & 40.9 & 72.9 \\
S$^2$IMI & 99.6* & 88.0 & 86.6 & 82.1 & 56.4 & 55.4 & 36.2 & 72.0 \\
Ours & \textbf{100.0*} & \textbf{92.3} & \textbf{90.9} & \textbf{83.7} & \textbf{81.1} & \textbf{81.5} & \textbf{69.8} & \textbf{85.6} \\ 
\bottomrule[1.2pt]

\textbf{Source: Inc-v4} & \multicolumn{7}{c}{Target model} & \multicolumn{1}{l}{} \\
 \cline{2-8}
Attack & Inc-v3 & Inc-v4 & IncRes-v2 & Res-101 & Inc-v3$_{ens3}$ & Inc-v3$_{ens4}$ & IncRes-v2$_{ens}$ & Avg. \\ 
\hline
SINI & 84.3 & \textbf{100.0*} & 76.1 & 69.8 & 39.9 & 37.7 & 24.3 & 61.7 \\
VTMI & 82.9 & 99.7* & 75.8 & 68.1 & 45.3 & 44.3 & 30.9 & 63.9 \\
Admix & 89.1 & 99.8* & 84.4 & 78.5 & 55.9 & 53.2 & 34.0 & 70.7 \\
GRA & 89.6 & 99.2* & 84.7 & 81.6 & 66.4 & 66.2 & 52.3 & 77.1 \\
S$^2$IMI & 91.5 & 99.4* & 86.3 & 81.4 & 57.3 & 55.8 & 37.2 & 72.7 \\
Ours & \textbf{94.5} & 99.7* & \textbf{91.2} & \textbf{83.7} & \textbf{79.3} & \textbf{79.6} & \textbf{68.9} & \textbf{85.3} \\

\bottomrule[1.2pt]
\textbf{Source: IncRes-v2} & \multicolumn{7}{c}{Target model} & \multicolumn{1}{l}{} \\
 \cline{2-8}
 Attack & Inc-v3 & Inc-v4 & IncRes-v2 & Res-101 & Inc-v3$_{ens3}$ & Inc-v3$_{ens4}$ & IncRes-v2$_{ens}$ & Avg. \\ \hline
SINI & 85.4 & 82.2 & 99.9* & 74.6 & 51.1 & 45.7 & 36.2 & 67.9 \\
VTMI & 81.8 & 78.7 & 98.5* & 71.6 & 51.1 & 45.9 & 40.8 & 66.9 \\
Admix & 91.6 & 90.0 & 99.9* & 86.0 & 69.3 & 62.5 & 52.1 & 78.8 \\
GRA & 88.4 & 87.0 & 98.6* & 81.5 & 71.4 & 68.4 & 65.3 & 80.1 \\
S$^2$IMI & 90.6 & 89.2 & 98.0* & 84.7 & 69.4 & 63.2 & 55.6 & 78.7 \\
Ours & \textbf{94.1} & \textbf{92.7} & \textbf{100.0*} & \textbf{86.8} & \textbf{82.7} & \textbf{82.0} & \textbf{79.7} & \textbf{88.3} \\ 
\bottomrule[1.2pt]
\textbf{Source: Res-101} & \multicolumn{7}{c}{Target model} & \multicolumn{1}{l}{} \\
 \cline{2-8}
 Attack & Inc-v3 & Inc-v4 & IncRes-v2 & Res-101 & Inc-v3$_{ens3}$ & Inc-v3$_{ens4}$ & IncRes-v2$_{ens}$ & Avg. \\  \hline
SINI & 78.4 & 73.2 & 72.6 & 99.9* & 45.0 & 42.7 & 29.0 & 63.0 \\
VTMI & 79.9 & 74.8 & 74.8 & 99.7* & 55.2 & 52.3 & 38.5 & 67.9 \\
Admix & 85.5 & 81.3 & 81.0 & \textbf{100.0*} & 58.8 & 55.6 & 38.1 & 71.5 \\
GRA & 87.8 & 82.8 & 84.0 & 99.7* & 70.7 & 70.8 & 58.1 & 79.1 \\
S$^2$IMI & 90.9 & 86.8 & 86.4 & 99.7* & 67.6 & 62.9 & 48.1 & 77.5 \\
Ours & \textbf{91.4} & \textbf{88.7} & \textbf{89.8} & 99.8* & \textbf{82.9} & \textbf{83.5} & \textbf{75.6} & \textbf{87.4} \\ 
\bottomrule[1.2pt]
\end{tabular}
% }
\end{table*}

\subsection{Experiment Setup}
\textbf{Dataset.} Following previous studies~\cite{long2022frequency,dong2018boosting}, we evaluate the performance of the proposed method on the widely used ImageNet-Compatible dataset\footnote{\url{https://github.com/cleverhans-lab/cleverhans/tree/master/cleverhans_v3.1.0/examples/nips17_adversarial_competition/dataset}}, which contains 1,000 images with the resolution of $299\times 299\times 3$.

\textbf{Models.} Following the GRA~\cite{zhu2023boosting}, we select four classical source models to craft adversarial examples, containing Inceptionv3 (Inc-v3), Inception-v4 (Inc-v4), Inception-Resnetv2 (IncRes-v2) and Resnet-v2-101 (Res-101). For evaluating the
attack performance, target models consist of the source models above and four adversarially trained models, including Inc-v3$_{adv}$~\cite{madry2017towards}, Inc-v3$_{ens3}$, Inc-v3$_{ens4}$, and IncRes-v2$_{ens}$~\cite{tramer2017ensemble}. Additionally, advanced defense methods that cover R\&P~\cite{xie2017mitigating}, NIPS-r3~\cite{thomas2017defense}, HGD~\cite{lin2020nesterov},  JPEG~\cite{guo2017countering}, (RS)~\cite{jia2019certified} and NRP~\cite{naseer2020self} are also taken into consideration. Finally, we select five transformer-based models as target models to expand our evaluation, namely ViT~\cite{dosovitskiy2020image}, LeViT~\cite{graham2021levit}, ConViT~\cite{d2021convit}, Twins~\cite{chu2021twins}, and PiT~\cite{heo2021rethinking}.

\textbf{Baseline Methods.}
To evaluate the performance of our proposed method, we conduct a comparative analysis with various state-of-the-art attack methods, including SINI-FGSM~\cite{lin2020nesterov}, VTMI-FGSM~\cite{wang2021enhancing}, Admix~\cite{wang2021admix}, GRA~\cite{zhu2023boosting}, and S$^2$IMI-FGSM~\cite{long2022frequency}. Additionally, we incorporate the combined transformation~\cite{wang2021enhancing} for compatibility verification, denoted as CT, which combines DI~\cite{xie2019improving}, TI~\cite{dong2019evading}, and SI~\cite{lin2020nesterov}. It's noteworthy that Admix inherently includes SI-FGSM~\cite{lin2020nesterov}. Consequently, Admix-CT and SINI-CT introduce only two additional augmentation transformations, namely DI and TI. For brevity in the subsequent context, VTMI-FGSM and S$^2$IMI-FGSM are referred to as VTMI and S$^2$IMI, respectively.

\textbf{Parameter Setting.} The attack settings align with previous works~\cite{lin2020nesterov,wang2021enhancing,wang2021admix,zhu2023boosting,long2022frequency}, where the maximum perturbation is $\epsilon=16$, the iteration is $T=10$, and the step size is $\alpha=\epsilon/T$. Specifically, we set the decay factor to $\mu=1.0$ for MI, the transformation probability to $p=0.5$ for DI, the kernel length to $k = 7$ for TI, and the number of copies to $m_1 = 5$ for SI. For VTMI, we set the hyper-parameter to $\beta = 1.5$, and the number of sampling examples is $N=20$. In the case of Admix, we set the number of copies to $m_1 = 5$, the sample number to $m_2 = 3$, and the admix ratio $\eta = 0.2$. For S$^{2}$I-FGSM, we set the tuning factor for $\mathcal{M}$ as $\rho = 0.5$, and the number of spectrum transformations as $N = 20$. Concerning GRA, we set the sample quantity as $N = 20$, the upper bound factor of the sample range $\beta$ is 3.5, and the attenuation factor $\eta$ is 0.94. For our proposed attacks, we set the sample quantity as $N = 10$. At each iteration, we randomly select a layer $k$ to perform feature mixing, and the weights of the feature mixing are also randomly determined.

\subsection{Attack with a Single Method}
In this section, we compare our method with several state-of-the-art attacks, including SINI~\cite{lin2020nesterov}, VTMI~\cite{wang2021enhancing}, Admix~\cite{wang2021admix}, GRA~\cite{zhu2023boosting}, and S$^2$IMI~\cite{long2022frequency} across four source models, and evaluate the attacks on seven target models. From Table \ref{tab:single}, it is evident that existing attack methods achieve commendable results on normally trained models but exhibit limited efficacy against defense models. In contrast, our method not only demonstrates superior performance across all normally trained models but also exhibits pronounced dominance over defense models. This effectiveness can be attributed to our innovative application of feature mixing and cross-frequency meta-optimization. Moreover, the average attack success rates on four source models of our method are the highest (Inc-v3: $85.6\%$,  Inc-v4: $85.3\%$, IncRes-v2:  $88.3\%$, and Res-101: $87.4\%$), significantly surpassing other attack methods.

\begin{table*}[]
\centering
\caption{The attack success rates (\%) on seven models by six gradient-based iterative attacks augmented with CT. The adversarial examples
are generated on Inc-v3, Inc-v4, IncRes-v2, and Res-101 separately. The adversarial examples are generated on Inc-v3, Inc-v4,
IncRes-v2, and Res-101 separately. $*$ denotes the success rate of the white-box attack and the result in bold is the best.}
\label{tab:ct}
% \resizebox{0.95\linewidth}{!}{
\begin{tabular}{lcccccccc}
\bottomrule[1.2pt]
\textbf{Source: Inc-v3} & \multicolumn{7}{c}{Target model} & \multicolumn{1}{l}{} \\
 \cline{2-8}
 Attack & Inc-v3 & Inc-v4 & IncRes-v2 & Res-101 & Inc-v3$_{ens3}$ & Inc-v3$_{ens4}$ & IncRes-v2$_{ens}$ & Avg. \\  \hline

SINI-CT & \textbf{100.0*} & 83.0 & 80.3 & 73.5 & 61.0 & 59.0 & 41.7 & 71.2 \\
VTMI-CT & \textbf{100.0*} & 86.2 & 82.1 & 76.9 & 72.5 & 70.2 & 58.0 & 78.0 \\
Admix-CT & \textbf{100.0*} & 89.7 & 87.9 & 82.8 & 76.4 & 73.9 & 60.0 & 81.5 \\
GRA-CT & 99.9* & 93.3 & 91.1 & 89.5 & 87.8 & 87.3 & 80.4 & 89.9 \\
S$^2$IMI-CT & 99.8* & 94.2 & 93.4 & 91.1 & 89.2 & 87.8 & 78.6 & 90.6 \\
Ours-CT & \textbf{100.0*} & \textbf{96.7} & \textbf{95.8} & \textbf{92.6} & \textbf{92.4} & \textbf{92.7} & \textbf{84.2} & \textbf{93.5} \\

\bottomrule[1.2pt]
\textbf{Source: Inc-v4} & \multicolumn{7}{c}{Target model} & \multicolumn{1}{l}{} \\
 \cline{2-8}
 Attack & Inc-v3 & Inc-v4 & IncRes-v2 & Res-101 & Inc-v3$_{ens3}$ & Inc-v3$_{ens4}$ & IncRes-v2$_{ens}$ & Avg. \\  \hline
 
SINI-CT & 86.9 & 99.9* & 83.7 & 74.9 & 66.4 & 62.5 & 51.4 & 75.1 \\
VTMI-CT & 88.6 & 99.5* & 84.2 & 77.9 & 74.0 & 72.0 & 61.9 & 79.7 \\
Admix-CT & 92.1 & 99.7* & 88.9 & 84.1 & 80.5 & 77.6 & 67.5 & 84.3 \\
GRA-CT & 93.5 & 99.5* & 91.5 & 87.0 & 88.3 & 87.3 & 81.4 & 89.8 \\
S$^2$IMI-CT & 95.6 & 99.2* & 93.8 & 90.9 & 88.2 & 86.8 & 81.1 & 90.8 \\
Ours-CT & \textbf{98.8} & \textbf{100.0*} & \textbf{96.9} & \textbf{92.8} & \textbf{92.7} & \textbf{93.3} & \textbf{88.3} & \textbf{94.7} \\

\bottomrule[1.2pt]
\textbf{Source: IncRes-v2} & \multicolumn{7}{c}{Target model} & \multicolumn{1}{l}{} \\
 \cline{2-8}
 Attack & Inc-v3 & Inc-v4 & IncRes-v2 & Res-101 & Inc-v3$_{ens3}$ & Inc-v3$_{ens4}$ & IncRes-v2$_{ens}$ & Avg. \\  \hline
 
SINI-CT & 89.8 & 87.7 & 99.8* & 84.6 & 75.6 & 70.6 & 65.7 & 82.0 \\
VTMI-CT &90.3	&89.5	&99.1	&86.2	&81.9	&77.6&	77.1&	86.0\\
Admix-CT & 93.5 & 92.0 & 99.4* & 89.7 & 87.4 & 84.1 & 81.0 & 89.6 \\
GRA-CT & 93.5 & 93.1 & 98.7* & 90.9 & 90.8 & 89.8 & 88.7 & 92.2 \\
S$^2$IMI-CT & 94.2 & 92.6 & 97.5* & 92.0 & 89.8 & 89.2 & 88.7 & 92.0 \\
Ours-CT & \textbf{97.4} & \textbf{97.0} & \textbf{100.0*} & \textbf{95.0} & \textbf{95.5} & \textbf{94.1} & \textbf{93.0} & \textbf{96.0} \\ 

\bottomrule[1.2pt]
\textbf{Source: Res-101} & \multicolumn{7}{c}{Target model} & \multicolumn{1}{l}{} \\
 \cline{2-8}
 Attack & Inc-v3 & Inc-v4 & IncRes-v2 & Res-101 & Inc-v3$_{ens3}$ & Inc-v3$_{ens4}$ & IncRes-v2$_{ens}$ & Avg. \\  \hline

 SINI-CT & 87.3 & 82.2 & 84.1 & 99.9* & 73.7 & 70.8 & 61.7 & 80.0 \\
 VTMI-CT & 87.0 & 84.4 & 84.7 & 99.9* & 81.1 & 77.6 & 71.4 & 83.7 \\
 Admix-CT & 91.9 & 87.9 & 88.8 & 99.9* & 84.5 & 81.5 & 73.7 & 86.9 \\
 GRA-CT & 91.3 & 86.8 & 89.5 & 99.8* & 88.9 & 88.3 & 85.4 & 90.0 \\
 S$^2$IMI-CT & 94.9 & 93.1 & 93.8 & 99.8* & 92.2 & 89.8 & 86.9 & 92.9 \\
 Ours-CT & \textbf{96.3} & \textbf{94.7} & \textbf{94.8} & \textbf{100.0*} & \textbf{93.8} & \textbf{93.6} & \textbf{89.2} & \textbf{94.6} \\
\bottomrule[1.2pt]
\end{tabular}
% }
\end{table*}

\subsection{Attack with the Combined Transformation}
In this part, we evaluated the performance of our proposed attack method by incorporating the Combined Transformation (CT) to enhance attack success rates and assess the compatibility. The CT is the combination of DI~\cite{xie2019improving}, TI~\cite{dong2019evading}, and SI~\cite{lin2020nesterov}.
The attack success rates are presented in Table~\ref{tab:ct}, and we can find that: \textbf{1)} Our method, when combined with CT, consistently achieves the highest attack success rates against various models compared to the five other attack methods. This indicates that our method exhibits strong compatibility with different transformations. \textbf{2)} Our method excels in attacking adversarially trained models compared to the existing five attack methods, validating findings observed without CT in Table~\ref{tab:single}. For instance, when Inc-v4 is used as the source model to generate adversarial examples for attacking IncRes-v2$_{ens}$, our method attains an impressive 88.3\% success rate, surpassing other methods such as GRA-CT (81.4\%) and S$^2$IMI-CT (81.1\%). This underscores the proficiency of our method not only in maintaining high success rates against standard models but also in significantly enhancing success rates against adversarially trained models.

\subsection{Attack with Ensemble Strategy}
We assess the effectiveness of our proposed method using the ensemble strategy introduced in~\cite{dong2018boosting}. Specifically, adversarial examples are crafted through an ensemble of Inc-v3, Inc-v4, and IncRes-v2 models. We then evaluate the attack success rates against ten different defense strategies, which include four adversarially trained models (Inc-v3$_{adv}$, Inc-v3$_{ens3}$, Inc-v3$_{ens4}$, IncRes-v2$_{ens}$) and six more advanced methods (HGD, R\&P, NIPS-r3, RS, JPEG, NRP). The results are outlined in Table~\ref{tab:ens}. From the table, it is clear that our attack method consistently attains the highest attack success rates against adversarial-trained models when compared with five other attack methods under the ensemble strategy, mirroring the results observed in Table~\ref{tab:single} and Table~\ref{tab:ct}. Additionally, our attack consistently achieves higher success rates when targeting other advanced defense methods except for a small gap on the NRP.

\begin{table*}[t]
\centering
\caption{The attack success rates (\%) on ten defended models attacked by adversarial examples crafted on Inc-v3, Inc-v4, and IncRes-v2 synchronously. The result in bold is the best.}
\label{tab:ens}
\begin{tabular}{lcccccccccc}
\bottomrule[1.2pt]
\textbf{Source: Ens} & \multicolumn{7}{c}{Target model} & \multicolumn{1}{l}{} \\
 \cline{2-11}
Attack & Inc-v3$_{adv}$ & Inc-v3$_{ens3}$ & Inc-v3$_{ens4}$ & IncRes-v2$_{ens}$ & JPEG & NRP & HGD & R\&P & NIPS-r3 & RS \\
\hline
 SINI & 63.7 & 66.5 & 63.3 & 45.3 & 84.9 & 36.1 & 47.8 & 46.3 & 58.2 & 41.1 \\
VTMI & 65.8 & 69.2 & 66.0 & 54.3 & 81.4 & 39.2 & 56.1 & 54.7 & 61.0 & 42.4 \\
Admix & 78.3 & 83.8 & 79.4 & 64.6 & 91.9 & 48.7 & 73.0 & 66.8 &76.3  & 48.3 \\
GRA & 86.2 & 85.0 & 83.7 & 78.0 & 90.7 & \textbf{69.7} & 76.7 & 77.7 & 82.0 & \textbf{66.7} \\
S$^2$IMI & 85.2 & 84.2 & 83.3 & 72.8 & 91.7 & 58.5 & 74.5 & 74.4 & 80.4 & 55.8 \\
Ours & \textbf{95.2} & \textbf{93.3} & \textbf{94.3} & \textbf{90.9} & \textbf{96.0} & 66.3 & \textbf{89.5} & \textbf{88.8} & \textbf{90.3} & 66.2\\ 
\bottomrule[1.2pt]
\end{tabular}
% }
\end{table*}

\begin{table}[ht!]
\centering
\caption{The attack success rates (\%) on five Transformer-based models by a single attack. The adversarial examples are generated on Inc-v3, Inc-v4,
IncRes-v2, and Res-101 separately. The result in bold is the best.}
\label{tab:vit}
\resizebox{0.99\linewidth}{!}{
\begin{tabular}{lccccccc}

\bottomrule[1.2pt]
\textbf{Source: Inc-v3} & \multicolumn{5}{c}{Target model} & \multicolumn{1}{l}{} \\
 \cline{2-6}
 Attack &  ViT & LeViT & ConViT & Twins & PiT & Avg. \\  \hline

SINI & 30.6 & 32.0 & 20.9 & 27.5 & 31.8 & 28.6 \\
VTMI & 32.5 & 39.2 & 27.8 & 34.9 & 38.7 & 34.6 \\
Admix & 34.7 & 34.7 & 29.0 & 34.2 & 39.1 & 34.3 \\
GRA & 40.7 & 51.7 & 37.1 & 43.3 & 49.8 & 44.5 \\
S$^2$IMI & 40.8 & 51.1 & 35.0 & 43.3 & 47.8 & 43.6 \\
Ours & \textbf{52.6} & \textbf{61.7} & \textbf{47.9} & \textbf{54.3} & \textbf{62.0} & \textbf{55.7} \\ 

\bottomrule[1.2pt]
\textbf{Source: Inc-v4} & \multicolumn{5}{c}{Target model} & \multicolumn{1}{l}{} \\
 \cline{2-6}
 Attack &  ViT & LeViT & ConViT & Twins & PiT & Avg. \\  \hline

SINI & 34.4 & 39.6 & 27.0 & 34.1 & 38.9 & 34.8 \\
VTMI & 35.4 & 46.1 & 32.5 & 42.4 & 41.3 & 39.5 \\
Admix & 40.0 & 49.8 & 34.1 & 45.5 & 47.5 & 43.4 \\
GRA & 44.9 & 61.3 & 42.9 & 53.6 & 58.9 & 52.3 \\
S$^2$IMI & 43.5 & 57.1 & 37.4 & 51.1 & 54.1 & 48.6 \\
Ours & \textbf{50.7} & \textbf{66.2} & \textbf{50.4} & \textbf{62.7} & \textbf{66.1} & \textbf{59.2} \\ 

\bottomrule[1.2pt]
\textbf{Source: IncRes-v2} & \multicolumn{5}{c}{Target model} & \multicolumn{1}{l}{} \\
 \cline{2-6}
 Attack &  ViT & LeViT & ConViT & Twins & PiT & Avg. \\  \hline
SINI & 36.5 & 41.7 & 30.4 & 34.5 & 40.8 & 36.8 \\
VTMI & 35.0 & 45.2 & 31.9 & 41.5 & 42.3 & 39.2 \\
Admix & 42.2 & 55.5 & 39.3 & 47.8 & 53.3 & 47.6 \\
GRA & 46.2 & 60.5 & 46.6 & 54.8 & 62.0 & 54.0 \\
S$^2$IMI & 43.4 & 58.2 & 43.5 & 52.3 & 57.3 & 50.9 \\
Ours & \textbf{52.1} & \textbf{62.4} & \textbf{51.5} & \textbf{58.2} & \textbf{65.7} & \textbf{58.0} \\ 

\bottomrule[1.2pt]
\textbf{Source: Res-101} & \multicolumn{5}{c}{Target model} & \multicolumn{1}{l}{} \\
 \cline{2-6}
 Attack &  ViT & LeViT & ConViT & Twins & PiT & Avg. \\  \hline
 
 SINI & 33.2 & 39.6 & 28.1 & 32.2 & 36.7 & 34.0 \\
VTMI & 36.6 & 45.6 & 31.7 & 42.1 & 42.7 & 39.7 \\
 Admix & 38.3 & 48.4 & 33.4 & 41.1 & 45.4 & 41.3 \\
 GRA & 48.0 & 55.1 & 40.3 & 49.1 & 53.0 & 49.1 \\
 S$^2$IMI & 42.9 & 58.0 & 37.6 & 47.6 & 54.3 & 48.1 \\
 Ours & \textbf{53.4} & \textbf{65.4} & \textbf{52.5} & \textbf{57.9} & \textbf{67.3} & \textbf{59.3}
 \\ \bottomrule[1.2pt]
\end{tabular}
}
\end{table}

\subsection{Attacking Transformer-based Models}
In this section, we extend our evaluation to include five transformer-based models, namely ViT~\cite{dosovitskiy2020image}, LeViT~\cite{graham2021levit}, ConVI~\cite{d2021convit}, Twins~\cite{chu2021twins}, and PiT~\cite{heo2021rethinking}. The results in Table~\ref{tab:vit} indicate that our proposed method consistently achieves higher attack success rates against transformer-based models compared to other attack methods. For instance, when considering Inc-v3 as the source model, our method achieves an average success rate of 55.7\%, outperforming other attack methods such as SINI (28.6\%), VTMI (34.6\%), Admix (34.3\%), GRA (44.5\%), and S$^2$IMI (43.6\%). Similar trends are observed across other source models (Inc-v4, IncRes-v2, and Res-101), showcasing the consistent superiority of our proposed method.

\subsection{Ablation Study}

%In this part, we conduct a series of ablation studies to evaluate the effectiveness of feature mixing and meta-learning in our proposed methods.

\begin{figure*}[ht!]
    \centering
    \begin{subfigure}{0.37\linewidth}
         \centering
\includegraphics[width=\textwidth]{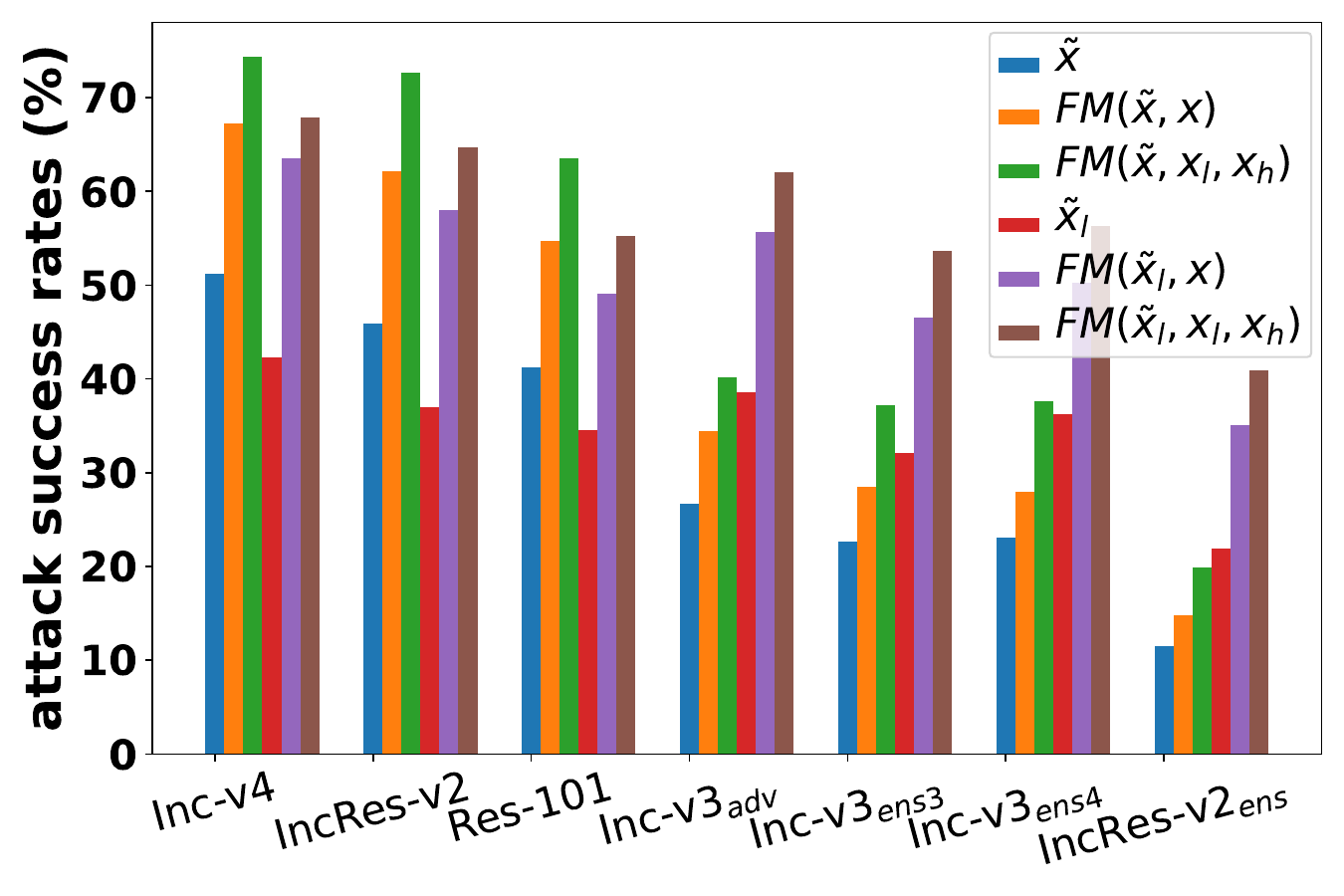}
\caption{Inc-v3}  \label{fig:ablation_incv3}
\end{subfigure}
    \begin{subfigure}{0.37\linewidth}
         \centering
\includegraphics[width=\textwidth]{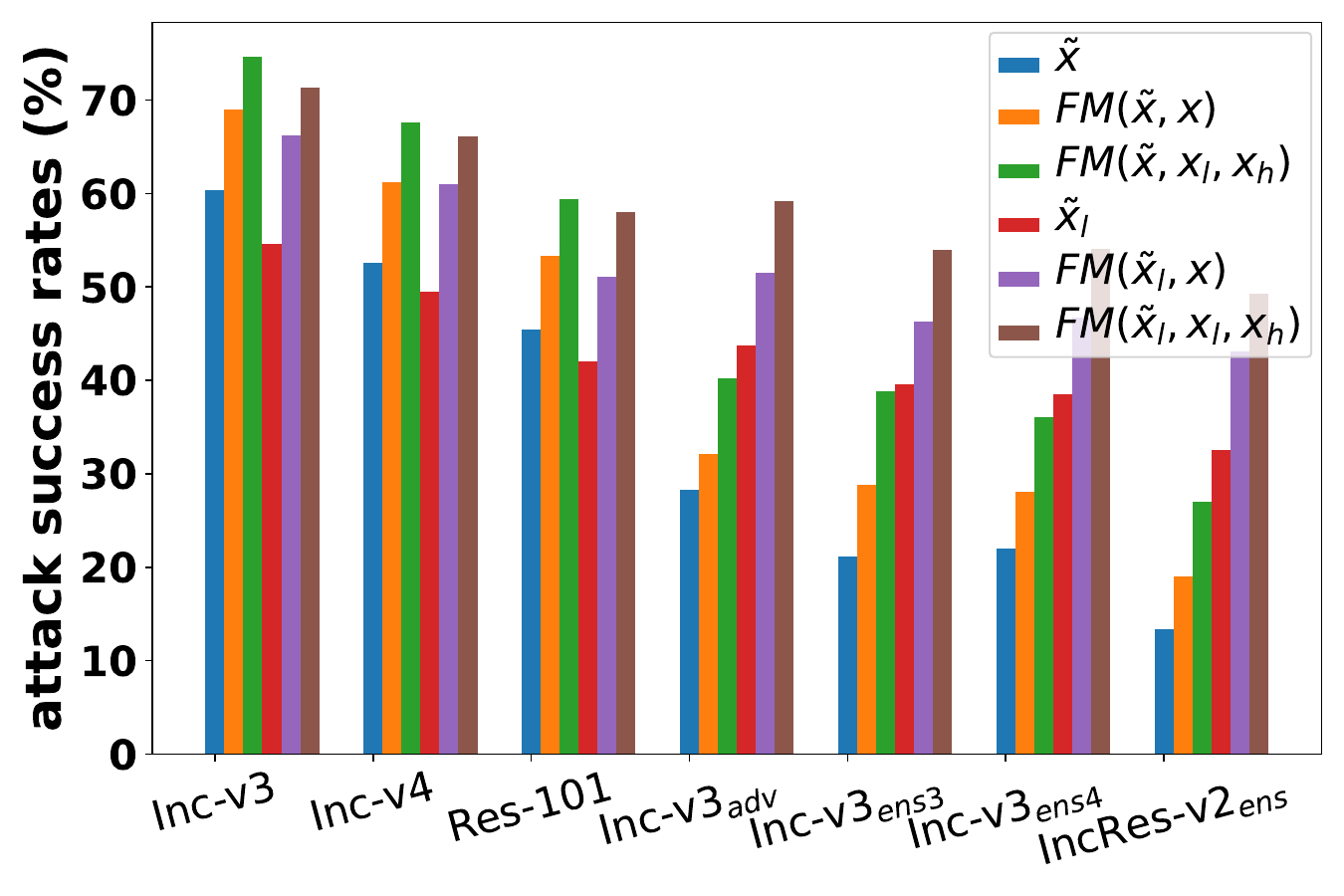}
\caption{IncRes-v2}  \label{fig:ablation_inres}
\end{subfigure}
    \caption{Comparison with different feature mixing strategies.}
    \label{fig:mix}
\end{figure*}

\noindent \textbf{(1) Different feature mixing strategies: }
In this part, we conduct a comparison between using different feature mixing strategies under the training framework without meta-optimization. The consider six feature mixing strategies are $\tilde{x}$, $FM(\tilde{x}, x)$, $FM(\tilde{x}, x_l, x_h)$, $\tilde{x}_l$, $FM(\tilde{x}_l, x)$, and $FM(\tilde{x}_l, x_l, x_h)$. From the results shown in Figure~\ref{fig:mix}, we can have the following observations: \textbf{1)} Comparing $\tilde{x}$ and $\tilde{x}_l$, we can find $\tilde{x}$ yields a higher attack success rate against regular CNNs, while $\tilde{x}_l$ exhibits higher transferability in attacking defense models. 
{These results further highlight the conflict issue between using $\tilde{x}$ and its corresponding low-frequency part $\tilde{x}_l$. }\textbf{2)} Mixing the features of the clean sample $x$ or its decomposed parts $(x_l, x_h)$ leads to a significant performance boost in attacking both regular CNNs and defense models. Moreover, $(x_l, x_h)$ achieves a higher performance than $x$.  These results indicate that performing feature mixing from both the low-frequency and high-frequency components of the clean sample can better explore the different frequency spectrum and improve transferability.

\begin{figure*}[ht!]
    \centering
    \begin{subfigure}{0.37\linewidth}
         \centering
\includegraphics[width=\textwidth]{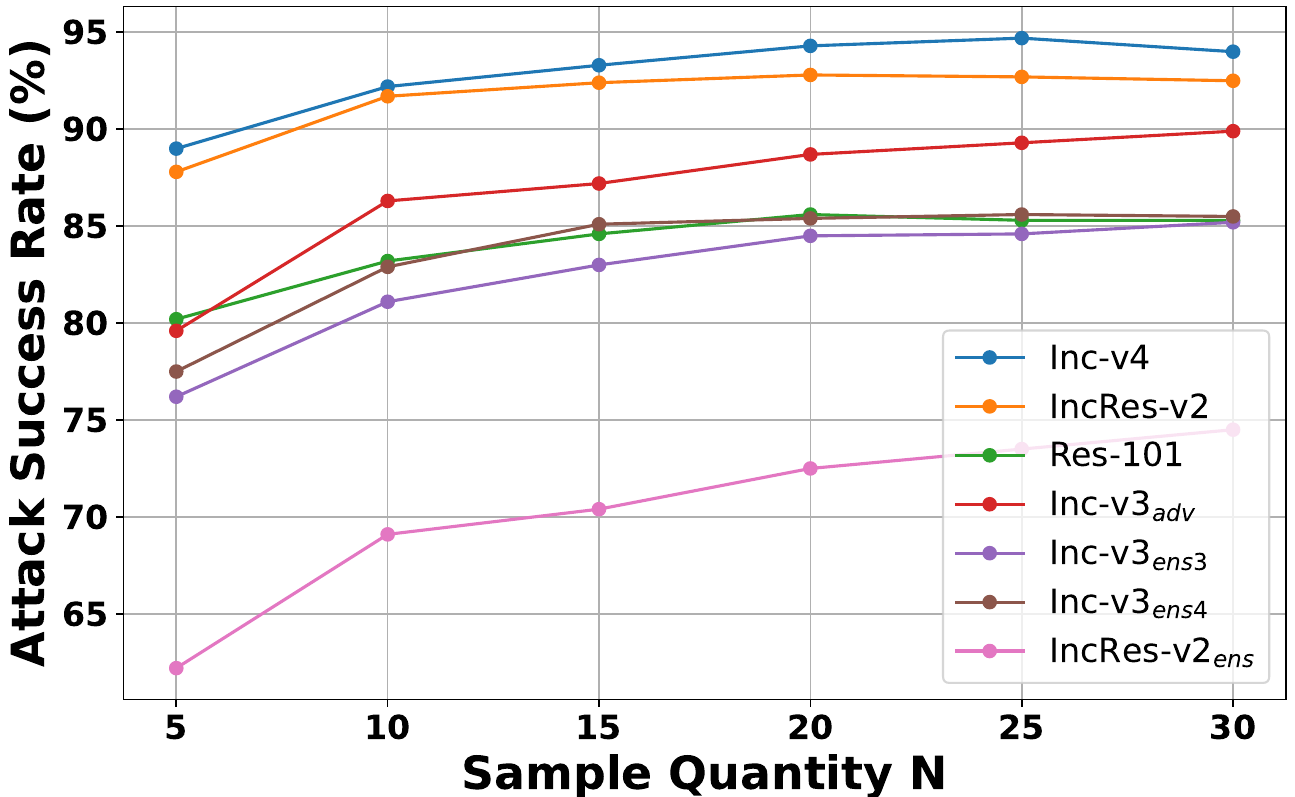}
\caption{Inc-v3}  \label{fig:n_incv3}
\end{subfigure}
    \begin{subfigure}{0.37\linewidth}
         \centering
\includegraphics[width=\textwidth]{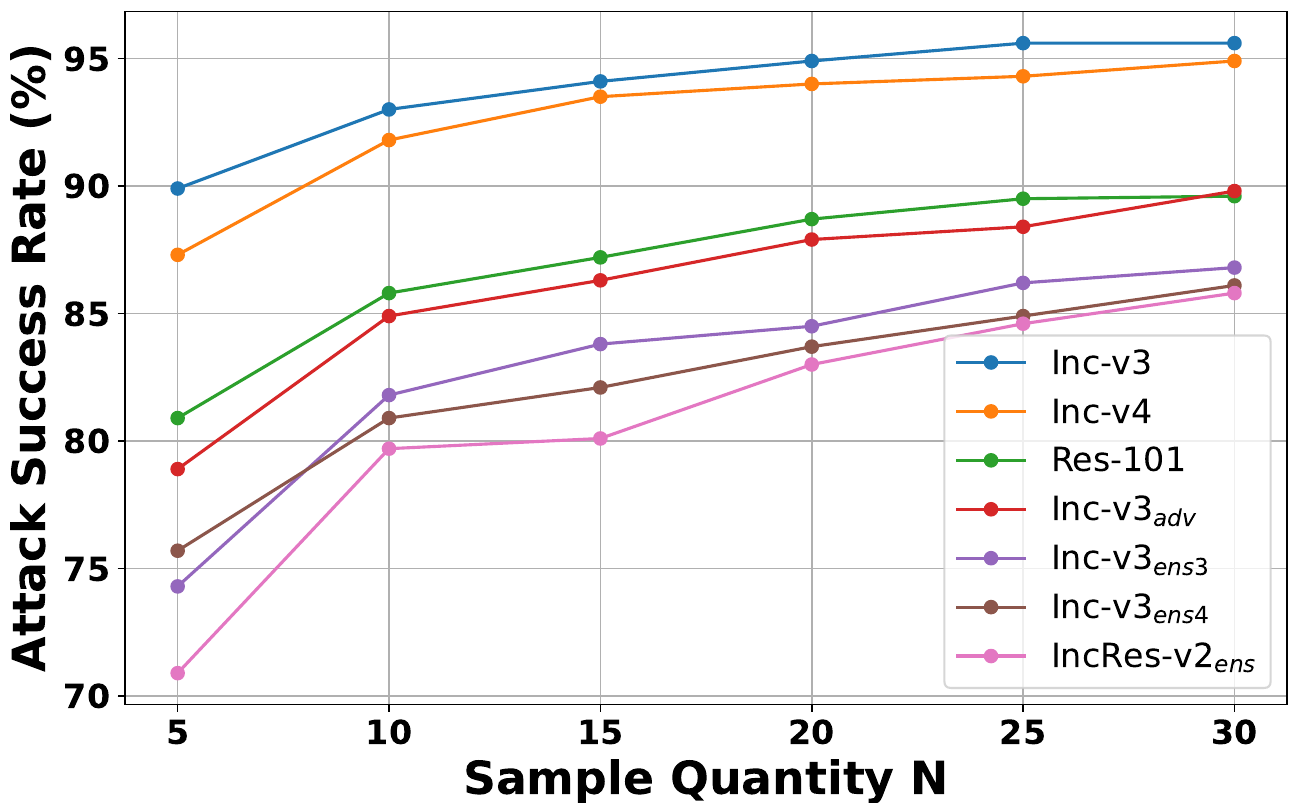}
\caption{IncRes-v2}  \label{fig:n_incres}
\end{subfigure}
    \caption{The influence of using different sample quantity $N$ in the cross-frequency meta optimization.}
    \label{fig:N}
\end{figure*}

\noindent \textbf{(2) Influence of sample quantity $N$ in meta optimization: }
The sample quantity $N$ in the cross-frequency meta-optimization is a hyper-parameter that will influence the inner gradients of both the meta-train and meta-test steps. To analyze its impact, we plotted the attack success rate in Figure \ref{fig:N} using Inc-v3 and IncRes-v2 as the source models to craft adversarial examples. From the figure, it can be observed that the attack success rate increases as $N$ increases. These results suggest that optimizing for more iterations could increase the transferability of the adversarial samples. To maintain a similar computation cost as GRA~\cite{zhu2023boosting} and S$^2$IMI~\cite{long2022frequency}, we set $N = 10$ in this study to allow for a fair comparison with these methods.

\noindent \textbf{(3) Different combinations in meta optimization: }
In Table \ref{tab:meta}, we present a comparative analysis of various feature combinations during meta-training and meta-testing stages, using Inc-v3 and IncRes-v2 as source models. The table reports the attack success rates against four normally-trained models and four adversarially-trained models.

From the table, we can find the following observations: 
\textbf{1)} In the 1st row (AFM, AFM), utilizing adversarial samples to guide both meta-train and meta-test optimization demonstrates superior performance against normally trained models but exhibits limited effectiveness when targeting adversarially-trained models.
\textbf{2)} Conversely, as shown in the 2nd row (LF-AFM, LF-AFM), employing low-frequency guidance during both meta-train and meta-test optimization results in superior performance when attacking adversarially trained models.
\textbf{3)} In the 3rd row (AFM, LF-AFM), we only observe a slight improvement when attacking adversarially trained models compared to (Adv, Adv).
\textbf{4)} In the 4th row (LF-AFM, AFM), when using low-frequency guidance for meta-train and adversarial examples for meta-test, the meta-test can compensate for the limited attack performance on normally trained models caused by low-frequency guidance for meta-train optimization, thereby improving overall performance. This approach achieves the highest performance and effectively addresses conflicts arising from the two feature-mixing scenarios.

Lastly, in the 5th row, we conduct an additional comparison by iteratively swapping the order of AFM and LF-AFM in the meta-train and meta-test steps. This approach results in a noticeable improvement in attacking normally trained models, while the transferability of targeting defense models falls between the (AFM, LF-AFM) strategy and (LF-AFM, AFM) strategy.

\begin{table*}[ht!]
\centering
\caption{Comparison of the average attack success rate between using different features in the meta optimization.}
\label{tab:meta}
\begin{tabular}{llccccccccc}
\bottomrule[1.2pt]
\multicolumn{2}{l}{\textbf{Source: Inc-v3}} & \multicolumn{8}{c}{Target Model} \\
\cline{3-10}
Meta-train & Meta-test & Inc-v3 & Inc-v4 & IncRes-v2 & Res-101 & Inc-v3$_{adv}$ & Inc-v3$_{ens3}$ & Inc-v3$_{ens4}$ & IncRes-v2$_{ens}$ \\
\hline
AFM & AFM & 100.0 & 92.9 & 92.8 & 87.1 & 69.7 & 67.0 & 64.6 & 42.6 \\
LF-AFM & LF-AFM & 100.0 & 89.7 & 87.5 & 80.4 & 84.3 & 81.2 & 80.6 & 68.6 \\
AFM & LF-AFM & 100.0 & 93.4 & 92.5 & 88.5 & 69.6 & 68.2 & 66.8 & 45.0 \\
LF-AFM & AFM & 100.0 & 92.3 & 90.9 & 83.7 & 85.8 & 81.1 & 81.5 & 69.8 \\
\multicolumn{2}{l}{iterative order swapping} & 100.0 & 93.6 & 92.4 & 86.0 & 74.5 & 74.1 & 75.4 & 54.7 \\

\bottomrule[1.2pt]
\multicolumn{2}{l}{\textbf{Source: IncRes-v2}} & \multicolumn{8}{c}{Target Model} \\
\cline{3-10}
Meta-train & Meta-test & Inc-v3 & Inc-v4 & IncRes-v2 & Res-101 & Inc-v3$_{adv}$ & Inc-v3$_{ens3}$ & Inc-v3$_{ens4}$ & IncRes-v2$_{ens}$ \\
\hline
AFM & AFM & 93.2 & 92 & 99.5 & 87.5 & 67.3 & 68.7 & 64.1 & 58.9 \\
LF-AFM & LF-AFM & 91.5 & 90.3 & 100.0 & 82.2 & 84.3 & 82.9 & 80.7 & 79.3 \\
AFM & LF-AFM & 93.5 & 92.1 & 99.6 & 87.4 & 67.3 & 72.2 & 67.5 & 60.3 \\
LF-AFM & AFM & 94.1 & 92.7 & 100.0 & 86.8 & 86.8 & 82.7 & 82.0 & 79.7 \\
\multicolumn{2}{l}{iterative order swapping} & 94.5 & 92.6 & 99.7 & 88.5 & 76.4 & 77.3 & 74.9 & 68.5 \\
\bottomrule[1.2pt]
\end{tabular}
\end{table*}

\begin{table}[h]
\centering
\caption{Comparing four metrics related with perceptual similarity by six attack approaches.}
\label{tab:ssim}
\begin{tabular}{lcccccc}
\bottomrule[1.2pt]
 & SINI & VTMI & Admix & GRA & S$^2$MI & \textbf{Ours} \\
 \hline
 $\ell_\infty$ & 0.0627 & 0.0627 & 0.0627 & 0.0627 & 0.0627 & 0.0627 \\
$\ell_2$ & 561.37 & 567.10 & 557.29 & 552.58 & 574.90 & 586.48 \\
LF & 313.83 & 313.49 & 311.42 & 319.67 & 319.52 & 569.66 \\
FID & 105.00 & 93.62 & 109.91 & 96.21 & 107.44 & 99.69 \\
\bottomrule[1.2pt]
\end{tabular}
\end{table}

\subsection{Comparing the computation time of different methods.}
We compare the computation times of different methods
for generating adversarial samples of 1,000 images using an RTX-3090 GPU. It is worth mentioning that we set N=10 in our study to ensure a comparable computation cost to GRA and S$^2$IMI. From Table~\ref{tab:time}, we can find that our method is slightly more efficient than these two approaches.

\begin{table}[ht!]
\centering
\caption{Comparing the computation times required to generate adversarial samples for 1,000 images using different methods.}
\label{tab:time}
\begin{tabular}{lcccccc}
\bottomrule[1.2pt]
         & SINI & VTMI & Admix & GRA & S$^2$IMI & \textbf{Ours} \\\hline
Time (s) & 190  & 762  & 492   & 753 & 741  & 714 \\ \bottomrule[1.2pt]
\end{tabular}
\end{table}

\subsection{Comparing the perceptual similarity metrics of different methods}
\label{sec:similarity-metrics}

For comparing the perceptual similarity metrics of different methods, we use four different metrics for approximating the perceptual similarity, including conventional average $\ell_2$ distortion, maximum perturbation intensity ($\ell_\infty$), Frechet Inception Distance (FID) \cite{heusel2017gans} and the average distortion of low-frequency components (LF) based on 2D Discrete Wavelet Transform (DWT) from \cite{luo2022frequency} ($LF = \frac{1}{N} \sum_{i=1}^{N} \left\| x_{l}^{(i)} - \tilde{x}_{l}^{(i)} \right\|_2$). 

From the Table~\ref{tab:ssim}, when using $\ell_\infty$ norm with $\epsilon=16$, our method exhibits the same results as others in $\ell_\infty$ and similar outcomes in $\ell_2$ and FID. However,  our distortion of low-frequency components (LF) is higher than others since we focus on attacking low-frequency images.

\subsection{Visualization}

\begin{figure*}[ht!]
    \centering
\includegraphics[width=0.97\textwidth]{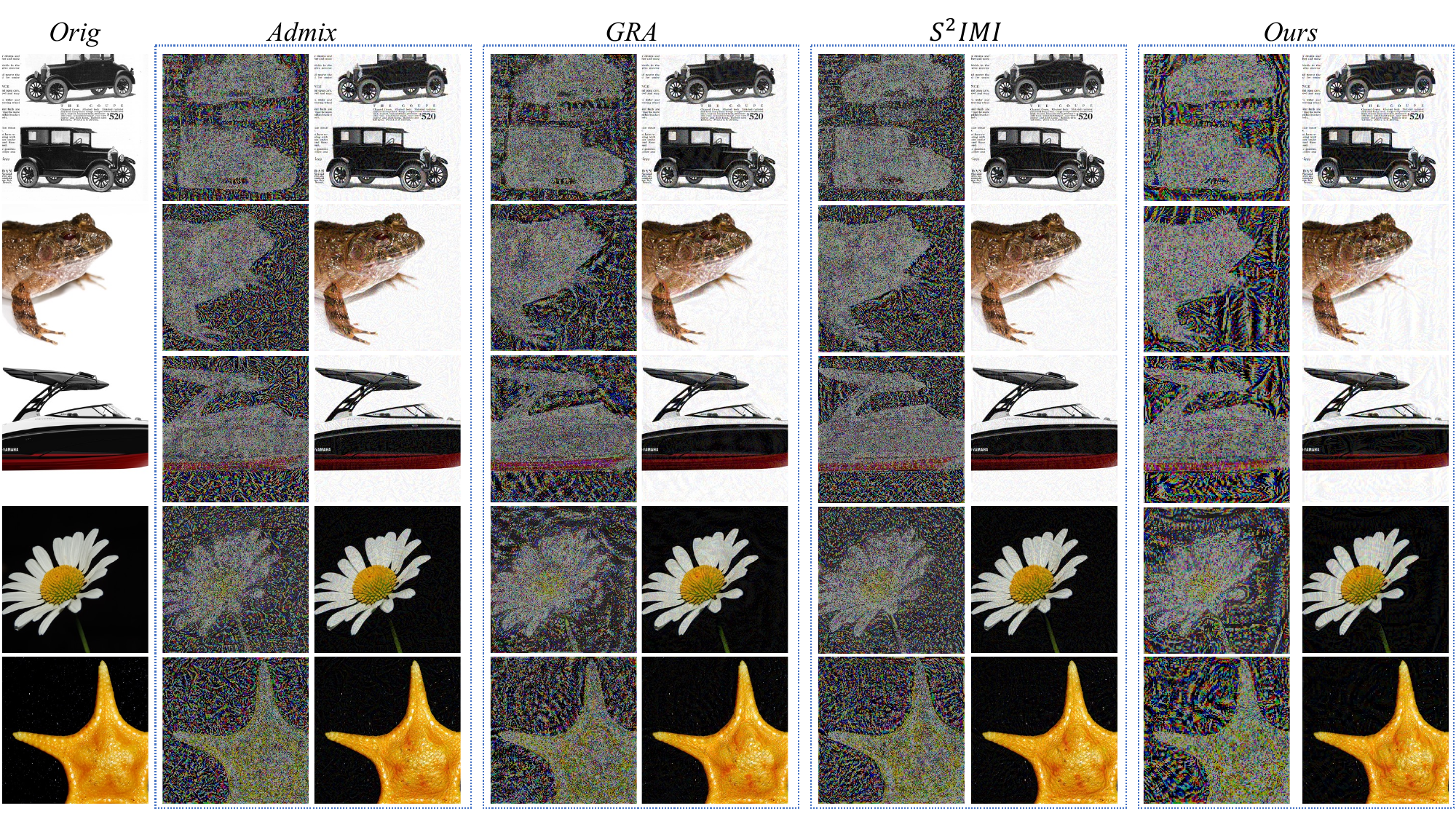}
    \caption{Visualization of adversarial images and the corresponding perturbations crafted by various attack methods.}
    \label{fig:adv-x}
\end{figure*}

In this section, we visualize some adversarial images and their corresponding perturbations generated by our method and other comparative approaches in Figure~\ref{fig:adv-x}. We observe that the adversarial noise produced by our method demonstrates a more pronounced impact in low-frequency background regions compared to other methods such as Admix and S$^2$IMI. Additionally, we note that the distortion of adversarial examples crafted by our method appears larger in low-frequency areas, consistent with the conclusion from the perceptual similarity metrics computed by the average distortion of the low-frequency component (LF) in Section~\ref{sec:similarity-metrics}.

\begin{figure*}[ht!]
    \centering
\includegraphics[width=0.85\textwidth]{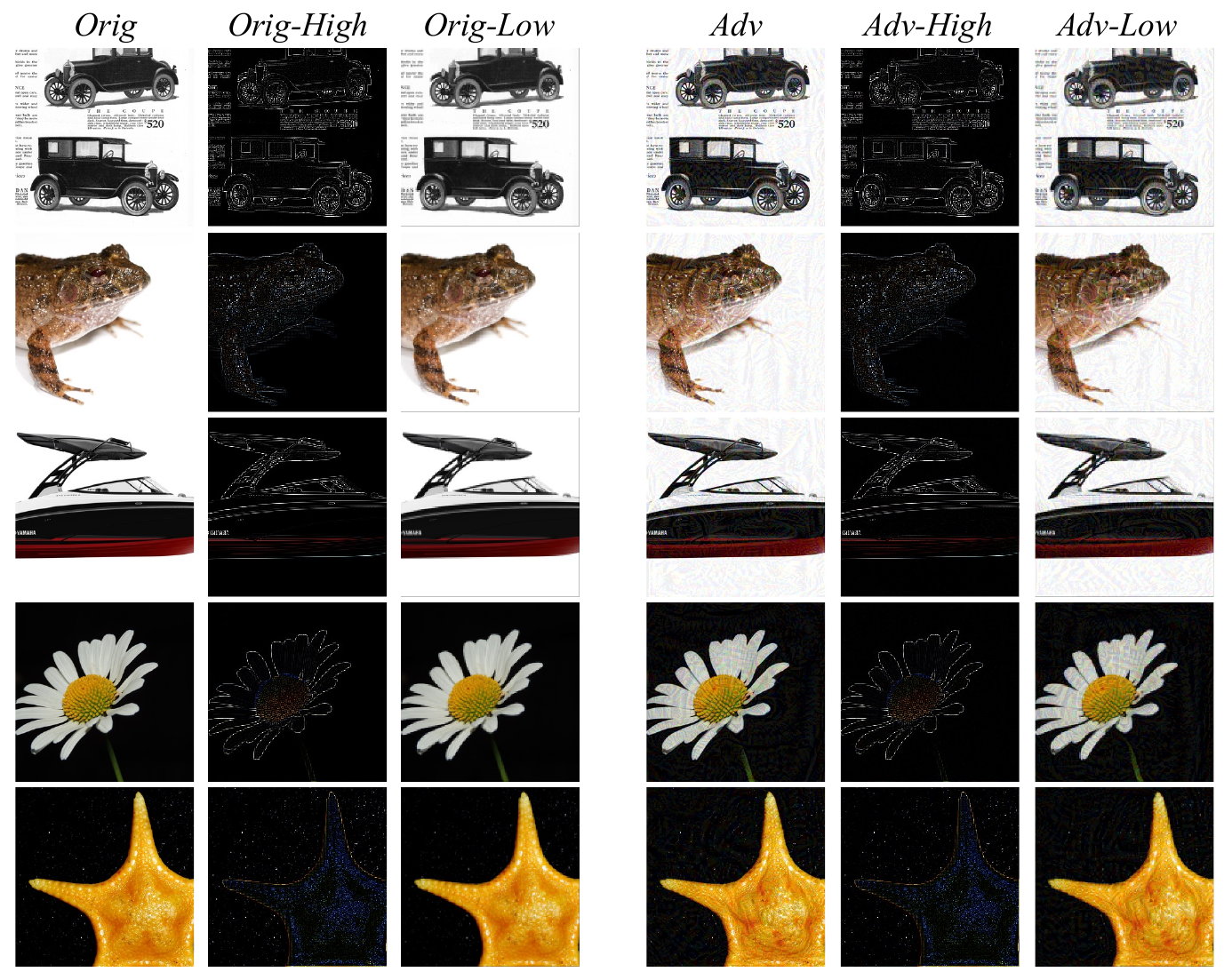}
    \caption{The frequency component of the original images and adversarial
images crafted by our methods.}
    \label{fig:adv-lf}
\end{figure*}

On the other hand, we visualize the high-frequency and low-frequency components of both original and adversarial images crafted by our method in Figure~\ref{fig:adv-lf}. Typically, the high-frequency component captures detailed textures, and we observe that the high-frequency components of the adversarial images closely resemble those of the original images. Moreover, when comparing the low-frequency component of adversarial examples with clean images, we note that adversarial examples are crafted by adding quasi-imperceptible perturbations primarily in the low-frequency component of the clean images.

\section{Conclusion}
In this paper, we introduced a frequency-based feature-mixing strategy and cross-frequency meta-learning framework for improving the disruptiveness of adversarial samples. We first found that leveraging the low-frequency parts can significantly increase the transferability of attacking defense models. However, a conflict arises when simultaneously utilizing adversarial samples and their corresponding low-frequency parts. Then, we further proposed a cross-frequency meta-learning framework to stabilize the gradients from these two parts. Experimental results validate the effectiveness of our methods in enhancing the transferability of attacks on both normally-trained and defense models.

\bibliographystyle{IEEEtran}
\bibliography{ref}

\end{document}